\documentclass[10pt,journal,compsoc]{IEEEtran}
\usepackage{amssymb}
\usepackage{multirow}
\usepackage{booktabs}
\usepackage{algorithm}
\usepackage{algorithmic}
\usepackage{subfigure}
\usepackage{flushend}
\usepackage{ragged2e}
\usepackage{url}
\usepackage{threeparttable}

\ifCLASSOPTIONcompsoc
  \usepackage[nocompress]{cite}
\else
  \usepackage{cite}
\fi

\ifCLASSINFOpdf
   \usepackage[pdftex]{graphicx}
\else
   \usepackage[dvips]{graphicx}
\fi

\usepackage{amsmath}
\usepackage{easyReview}
\title{
Pre-training General Trajectory Embeddings with Maximum Multi-view Entropy Coding
}

\author{
Yan~Lin,
Huaiyu~Wan, 
Shengnan~Guo,
Jilin~Hu,
Christian S. Jensen,
Youfang~Lin
\thanks{Corresponding author: Shengnan~Guo}
\IEEEcompsocitemizethanks{\IEEEcompsocthanksitem Yan Lin, Huaiyu Wan, Shengnan Guo and Youfang Lin are with the Beijing Key Laboratory of Traffic Data Analysis and Mining, School of Computer and Information Technoloty, Beijing Jiaotong University, Beijing 100044, China, and the Key Laboratory of Intelligent Passenger Service of Civil Aviation, CAAC, Beijing, 101318, China.
Jilin Hu and Christian S. Jensen are with the Department of Computer Science, Aalborg University, Aalborg, 9220, Denmark.
\protect\\
E-mail: ylincs@bjtu.edu.cn; 
hywan@bjtu.edu.cn; 
hujilin@cs.aau.dk; 
guoshn@bjtu.edu.cn; 
csj@cs.aau.dk;
yflin@bjtu.edu.cn.
}
\thanks{Manuscript received xx xx, xxxx; revised xx xx, xxxx.}
}

\newtheorem{definition}{Definition}
\newcommand{\paratitle}[1]{\vspace{1ex}\noindent \textbf{#1}}

\usepackage{color}

\begin{document}
\setreviewsoff

\IEEEtitleabstractindextext{%
\begin{abstract}
\justifying
Spatio-temporal trajectories provide valuable information about movement and travel behavior, enabling various downstream tasks that in turn power real-world applications. Learning trajectory embeddings can improve task performance but may incur high computational costs and face limited training data availability. Pre-training learns generic embeddings by means of specially constructed pretext tasks that enable learning from unlabeled data. Existing pre-training methods face (i) difficulties in learning general embeddings due to biases towards certain downstream tasks incurred by the pretext tasks, (ii) limitations in capturing both travel semantics and spatio-temporal correlations, and (iii) the complexity of long, irregularly sampled trajectories.

To tackle these challenges, we propose Maximum Multi-view Trajectory Entropy Coding (MMTEC) for learning general and comprehensive trajectory embeddings. We introduce a pretext task that reduces biases in pre-trained trajectory embeddings, yielding embeddings that are useful for a wide variety of downstream tasks. We also propose an attention-based discrete encoder and a NeuralCDE-based continuous encoder that extract and represent travel behavior and continuous spatio-temporal correlations from trajectories in embeddings, respectively. Extensive experiments on two real-world datasets and three downstream tasks offer insight into the design properties of our proposal and indicate that it is capable of outperforming existing trajectory embedding methods.
\end{abstract}

}

\maketitle
\IEEEdisplaynontitleabstractindextext
\IEEEpeerreviewmaketitle

\IEEEraisesectionheading{\section{Introduction}\label{sec:introduction}}
\IEEEPARstart {A} spatio-temporal trajectory is a long sequence of irregularly sampled (location, time) pairs that records the movement and travel behavior of a vehicle or a person.
With the widespread availability of GPS-equipped devices and the growing popularity of location-based services, trajectory data from sources such as in-vehicle navigation devices and smartphones have become increasingly available. 
This wealth of data enables a wide range of spatio-temporal trajectory mining tasks, including predicting future trajectories~\cite{kong2018hst}, estimating time of arrival~\cite{li2019learning,hong2020heteta,jin2021hierarchical}, anomaly detection~\cite{han2022deeptea}, and trajectory clustering~\cite{yao2017trajectory,liu2020online}. 
It is attractive to let such downstream tasks operate on embeddings of trajectories---d-dimensional vectors that capture essential aspects of trajectories---rather than on the "native" trajectories themselves. However, when doing so, the accuracy and comprehensiveness of the embeddings are crucial to achieving good performance.

\alert{
Existing trajectory mining methods that use embeddings commonly adopt an end-to-end embedding approach. This involves training a trajectory encoder alongside other components within the model, using task-specific learning objectives~\cite{zhao2019go,DBLP:journals/expert/SunGXZLFL22,DBLP:conf/wsdm/DangWPZ0C022}.
While intuitive and straightforward to implement, this approach also possesses limitations related to generalizability, efficiency, and real-world feasibility~\cite{DBLP:conf/iclr/AshukhaLMV20,DBLP:conf/icml/IshidaYS0S20}. Thus, the embeddings learned using this approach are tailored to specific tasks, rendering them less applicable across different tasks. The resulting need to retrain embeddings for each new task reduces computational efficiency. Moreover, the end-to-end embedding approach relies on the availability of extensive task labels to achieve satisfactory outcomes. Obtaining such labels can incur substantial time and financial costs.
Instead, we advocate focusing on pre-training general trajectory embedding methods. Such methods entail acquiring foundational embeddings through pretext tasks. This approach offers key advantages.
First, it enables utilization of existing unlabeled trajectory data, facilitating the capture of general spatio-temporal information inherent in trajectories. Consequently, the accuracy of predictions and generalization performance in downstream tasks can be enhanced~\cite{devlin2018bert,DBLP:conf/nips/YangDYCSL19,tian2020contrastive}.
Second, pre-trained embeddings can be fine-tuned with limited labeled data or can be applied directly to unsupervised tasks. This expedites downstream task training and improves computational efficiency~\cite{oord2018representation,bachman2019learning,DBLP:journals/corr/abs-2003-08271}.
}

\begin{figure*}[t]
    \centering
    \includegraphics[width=1.0\linewidth]{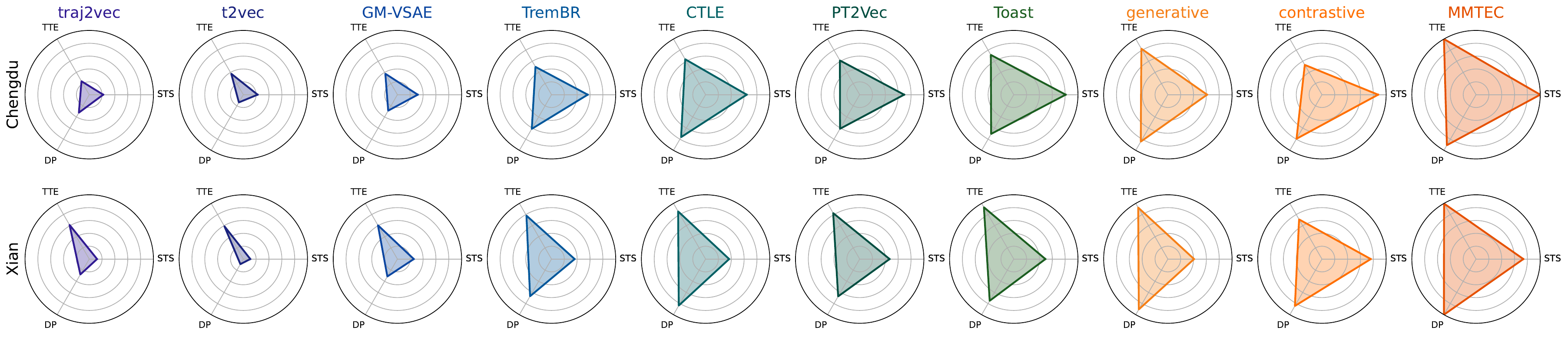}
    \caption{\alert{Comparison of the performance of trajectory embeddings on two datasets and three downstream tasks.}
    The axes represent the relative performance of an embedding on a downstream task. 
    STS, TTE, and DP denote the normalized Acc@1 for similar trajectory search, 100\% - MAPE for travel time estimation, and Acc@1 for destination prediction, respectively. 
    The area of a polygon represents the generality of a trajectory embedding method.
    }
    \label{fig:radar}
\end{figure*}

The design of a self-supervised pretext task for guiding the training of a trajectory encoder is central to the pre-training of general trajectory embeddings. The resulting process facilitates the extraction of universal insights from unlabeled trajectory data, encompassing elements of travel semantics such as purpose, destination, and route preference, alongside spatio-temporal movement patterns like speed and direction. Consequently, the resulting embeddings empower downstream analysis tasks that include prediction or classification.
\alert{
Thus, researchers are increasingly focusing on the pre-training of spatio-temporal trajectory embeddings.
However, three challenges remain unsolved.
}

\noindent
\alert{(1)~\textbf{
Reduced generalizability due to biases introduced into the learned embeddings.
}}
\alert{
To learn general embeddings of spatio-temporal trajectories that can enhance prediction accuracy and accelerate convergence across diverse downstream tasks, self-supervised pretext tasks must avoid introducing biases that favor particular types of downstream tasks.
In constrast, current pre-training methods employ pretext tasks that yield improved performance on certain downstream tasks, while causing reduced performance on others, as exemplified in Figure~\ref{fig:radar}.
}

\alert{
In particular, the proposals t2vec~\cite{li2018deep}, traj2vec~\cite{yao2017trajectory}, GM-VSAE~\cite{liu2020online}, TremBR~\cite{fu2020trembr}, and PT2Vec~\cite{li2023trajectory} employ a generative auto-regressive-based pretext task, mapping a trajectory into a latent space and subsequently reconstructing its original form from this latent space using an auto-regressive approach. While this pretext task aims at capturing intricate trajectory detail~\cite{oord2018representation}, it can inadvertently introduce biases that favor point-level tasks such as travel-time estimation and destination prediction.
Also, Toast~\cite{chen2021robust} explores the idea of pre-training trajectory embeddings through sequence discrimination~\cite{tian2020contrastive,chen2021robust}. However, this approach may introduce biases toward sequence-level tasks like similar trajectory search and trajectory-user linking.
}

\begin{figure}[t]
    \centering
    \subfigure[\alert{Travel semantics}] {
        \begin{minipage}[t]{0.72\linewidth}
        \centering
        \includegraphics[width=1.0\linewidth]{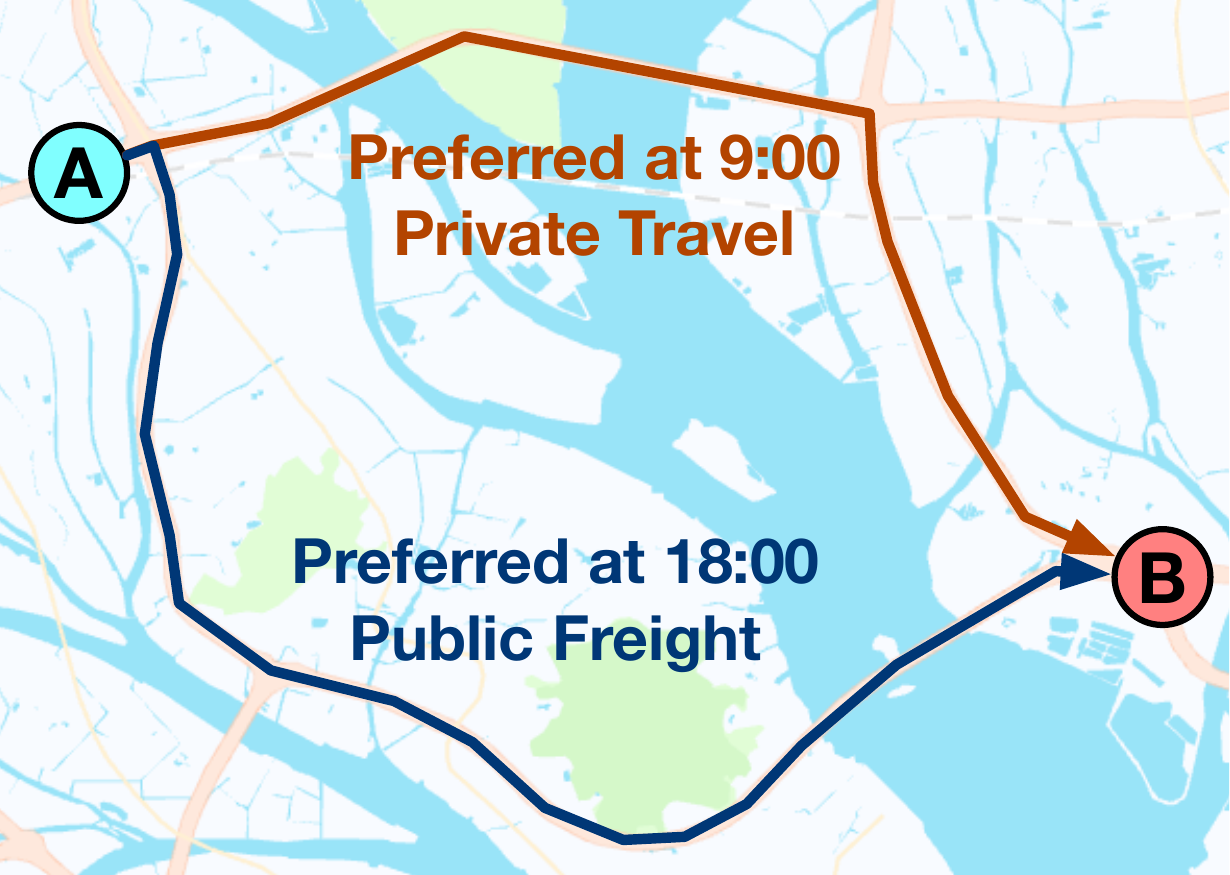}
        \end{minipage}
        \label{fig:aspects-travel-semantics}
    }
    \subfigure[\alert{Continuous ST correlations}] {
        \begin{minipage}[t]{0.21\linewidth}
        \centering
        \includegraphics[width=1.0\linewidth]{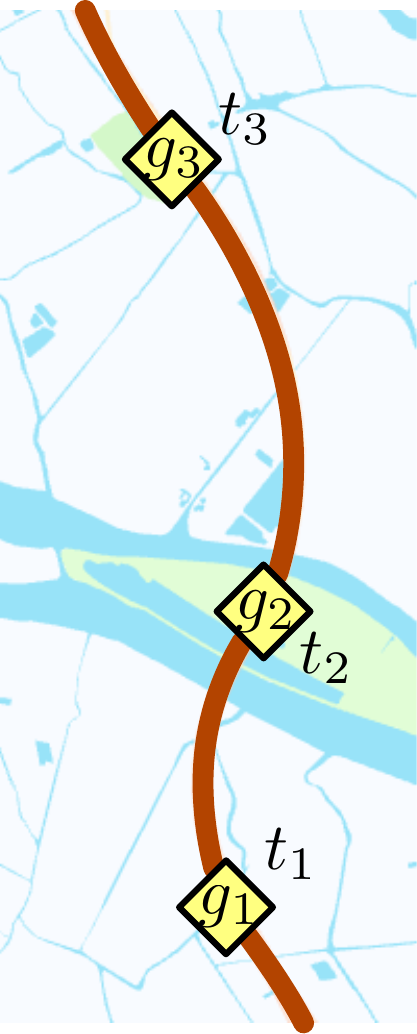}
        \end{minipage}
        \label{fig:aspects-continuous-st-correlations}
    }
    \caption{\alert{Two aspects of information embedded in trajectories.}}
    \label{fig:two-aspects-of-information}
\end{figure}

\noindent
\alert{(2)~\textbf{
Incomplete capture of informational aspects of trajectories.
}}
\alert{
To learn trajectory embeddings conducive to enhancing downstream task performance, it is paramount to capture a multitude of informational aspects of trajectories.
These aspects can be categorized as either travel semantics and continuous spatio-temporal correlations. Trajectories inherently capture movement in road networks and possess intrinsic semantic attributes that encapsulate travel intentions and route preferences~\cite{chen2021robust}, as exemplified in Figure~\ref{fig:aspects-travel-semantics}. 
Furthermore, trajectories represent continuous movements across time spans, as exemplified in Figure~\ref{fig:aspects-continuous-st-correlations}. 
Yet, existing trajectory embedding methods predominantly focus on one of these aspects, reducing the ability of the learned embeddings to capture comprehensively the aspects of trajectory information.
}

\alert{
More specifically, t2vec~\cite{li2018deep}, traj2vec~\cite{yao2017trajectory}, and GM-VSAE~\cite{liu2020online} prioritize modeling spatio-temporal correlations in trajectories, while disregarding travel semantics derived from the underlying road networks.
Next, TremBR~\cite{fu2020trembr}, PT2Vec~\cite{li2023trajectory}, and Toast~\cite{chen2021robust} incorporate travel semantics when embedding road network-constrained trajectories, while their modeling of spatio-temporal correlations is comparatively less developed.
}

\noindent
\alert{(3)~\textbf{
Difficulties at modeling long and irregularly sampled trajectories.
}}
\alert{
Long trajectories pose efficiency challenges to sequential models, and temporal irregularity in trajectories render it difficult to capture continuous spatio-temporal correlations.
}

\alert{
Existing trajectory embedding methods frequently draw inspiration from Natural Language Processing (NLP) and time series modeling, notably leveraging sequential models. 
Examples include t2vec~\cite{li2018deep}, traj2vec~\cite{yao2017trajectory}, GM-VSAE~\cite{liu2020online}, TremBR~\cite{fu2020trembr}, and PT2Vec~\cite{li2023trajectory} that all construct their trajectory encoders based on RNNs~\cite{hochreiter1997long}. Additionally, Toast~\cite{chen2021robust} utilizes a Transformer-based encoder~\cite{vaswani2017attention}.
However, RNN-based encoders suffer from the vanishing gradient issue~\cite{linzen2016assessing}, while Transformer-based encoders have quadratic time and memory complexity, rendering them suboptimal for capturing long-term correlations in trajectories. 
Conversely, both RNN and Transformer encoders solely update their hidden states upon receiving new input records, limiting their ability to capture the continuous dynamics characterizing trajectories~\cite{liang2021modeling}.
}

\alert{
To address the above challenges, we propose a novel trajectory embedding method called \textit{\underline{M}aximum \underline{M}ulti-view \underline{T}rajectory \underline{E}ntropy \underline{C}oding} (MMTEC). This method tackles the challenges as follows: 
(1) It leverages the principle of maximum entropy to learn unbiased, universal trajectory embeddings that benefits a wide range of downstream tasks; 
(2) It incorporates multiple views of trajectories into the maximum entropy loss, capturing both aspects of the information embedded within trajectories; 
(3) It encompasses an attention-based discrete trajectory encoder featuring globally-shared anchors in order to capture travel semantics and a continuous trajectory encoder grounded in NeuralCDE to extract continuous spatio-temporal trajectory dynamics.
}

In summary, the paper's main contributions are as follows:
\begin{itemize}
    \item We propose MMTEC, a method for pre-training spatio-temporal trajectory embeddings that learns general trajectory embeddings beneficial for the performance of various downstream tasks.
    \item We develop a trajectory multi-view scheme that incorporates both semantic and continuous spatio-temporal information into the learned embeddings.
    \item We build an attention-based discrete trajectory encoder and a NeuralCDE-based continuous trajectory encoder that can model different aspects of spatio-temporal correlations in trajectories comprehensively and efficiently.
    \item We report on an experimental study involving two real-world trajectory datasets and three downstream tasks, finding that MMTEC embeddings generalize well and improve downstream tasks' performance.
\end{itemize}

The remainder of the paper is organized as follows: 
Section~\ref{sec:related-work} reviews related work. 
Section~\ref{sec:preliminaries} covers the problem statement, definitions, and basic maximum entropy coding framework. 
Section~\ref{sec:methodology} details the implementation of our proposed method. 
Section~\ref{sec:experiments} reports on the experimental study, 
and Section~\ref{sec:conclusion} concludes the paper.

\section{Related Work}\label{sec:related-work}
\subsection{Learning Embeddings for Spatio-Temporal Data Mining}
Embedding learning is an inherent aspect of neural network data mining, as most neural network models are feature-based and require the input targets to be represented by latent embedding tensors. 
In spatio-temporal data mining, embedding learning is also crucial. Spatio-temporal objects, such as locations, road networks, and trajectories, often contain complex spatio-temporal information while being geographically and dynamically correlated with other objects. Capturing and fusing such information into embeddings is challenging.

Locations are fundamental units in spatio-temporal datasets, and location embedding techniques are used widely in spatio-temporal data mining models. 
A straightforward idea is to use an index-fetching embedding layer~\cite{kong2018hst,zhao2019go}, as locations such as POIs and road segments are often represented by discrete indices. 
However, this kind of embedding is trained through task-specific supervision, cannot properly capture spatio-temporal correlations between locations comprehensively, and often requires a large labeled dataset to work well.

To tackle the above problems, several methods pre-train embeddings of POIs utilizing large, unlabeled datasets. For example, DeepMove~\cite{zhou2018deepmove} proposes word2vec~\cite{Mikolov2013Efficient} that works on unlabeled movement data, while POI2Vec~\cite{feng2017poi2vec} and TALE~\cite{wan2021pre} incorporate spatial and temporal correlations between locations into the embedding vectors. 
Next, CTLE~\cite{lin2021pre} applies the idea of contextual embedding from language models to location embedding. These pre-training methods support downstream tasks such as location recommendation~\cite{feng2015personalized,zhou2018atrank} and location classification~\cite{yao2018representing,shimizu2020learning}.

Road segments and intersections in road networks can also be seen as locations. They contain complex information inherited from the topological structure of the road network. Using only an index-fetching embedding layer for road network representation will eliminate the important topological information. Therefore, methods such as IRN2Vec~\cite{wang2019learning}, Toast~\cite{chen2021robust}, and HRNR~\cite{wu2020learning} incorporate topological correlations in road segment embedding by coupling random walks on road networks with word2vec, by utilizing real-world trajectories to fuse network topology with traffic patterns, and by proposing a hierarchical graph neural network and an auto-encoding pre-training objective to learn more comprehensive embeddings of road segments.

In this paper, we focus on the pre-training of trajectory embeddings, thus emphasizing the travel semantics and spatio-temporal movements in a trajectory sequence as a whole.

\subsection{Pre-training Trajectory Embeddings}
The quality of trajectory embeddings is crucial for achieving good performance of trajectory analysis tasks. 
The most widely applied trajectory embedding methodology is to construct an end-to-end sequence encoder for generating latent representations~\cite{kong2018hst,zhao2019go}. 
However, the resulting trajectory embeddings often do not generalize and are difficult to migrate to other models or tasks. 
This methodology also assumed that substantial labeled training data is available, which may not be the case for some downstream tasks.

In part to address these issues, there is a growing interest in the pre-training of trajectory embeddings with self-supervised learning objectives.
\alert{
The notion of pre-training embeddings via self-supervised objectives stems from Language Modeling (LM)~\cite{devlin2018bert,DBLP:conf/acl/DuQLDQY022} and Computer Vision (CV)~\cite{oord2018representation,tian2020contrastive}, domains that embrace self-supervised pre-training techniques.
}
For instance, t2vec~\cite{li2018deep} builds an RNN-based trajectory encoder and applies it to GPS sequences to infer latent representations of trajectories based on sequential correlations of coordinates in the trajectories. 
GM-VSAE~\cite{liu2020online} further maps trajectories into a Gaussian latent space, so that the distributions of trajectories can be modeled. 
Traj2vec~\cite{yao2017trajectory} considers relative spatio-temporal properties, using the differences in a set of attributes between consecutive trajectory points as input features. 
TremBR~\cite{fu2020trembr} models travel semantics by mapping trajectories onto road networks and additionally incorporates temporal information by concatenating the travel time of each point with its latent embedding vector.
\alert{PT2Vec~\cite{li2023trajectory} partitions road networks into sub-networks to model map-matched trajectories more effectively.}

The above methods build pretext tasks based on generative objectives, including auto-encoding~\cite{hinton2006reducing} and auto-regression~\cite{pauls2011faster}, where a decoder is trained to recover the trajectories given their embeddings. Since the generative objectives focus on modeling the detailed sequential features, the learned embeddings are usually biased towards point-level tasks, such as travel time estimation.

Recently, Toast~\cite{chen2021robust} proposed a trajectory embedding pre-training method that combines a masked language model (MLM)~\cite{devlin2018bert} pretext task with a sequence discrimination contrastive pretext objective~\cite{tian2020contrastive,devlin2018bert}. Trajectory embeddings learned by contrastive objectives tend to be biased towards sequence-level tasks, such as trajectory-user linking.
It is evident that existing pretext tasks are inevitably biased towards certain types of downstream tasks, which conflicts with the goal of learning general trajectory embeddings that performs optimally across diverse tasks.

\alert{
We note that several existing methods were initially tailored to address specific tasks within their original contexts.
However, these methods are united by their underlying adherence to self-supervised pre-training paradigms, inherently enabling them to be used in other downstream tasks.
This paper's coverage is not confined to the initial application scopes of these related studies. Instead, we focus on the technical frameworks that form the core of these methods.
A comparison of three attributes of the methods is provided in Table~\ref{table:related-works-comparison}.
}

\section{Preliminaries}\label{sec:preliminaries}
\subsection{Definitions}

\begin{definition}
[Trajectory]
A trajectory is a sequence of timestamped GPS points $\mathcal T=\langle (g_1,t_1), (g_2,t_2), \dots,$ $(g_{|\mathcal T|},t_{|\mathcal T|}) \rangle$, where $g_i=(\mathrm{lng}_i, \mathrm{lat}_i)$ denotes the $i$-th GPS point, and $|\mathcal T|$ is the total number of GPS points in the trajectory.
\end{definition}

\begin{definition}
[Map-matched Trajectory]
A map-matched trajectory is a sequence of timestamped road segments $\mathcal R=\langle (s_1,t_1), (s_2,t_2), \dots, (s_{|\mathcal R|},t_{|\mathcal R|}) \rangle$, where $s_i\in \mathbb S$ denotes the $i$-th road segment, $\mathbb S$ is the set of all road segments in the area of interest, and $|\mathcal R|$ is the total number of road segments in the trajectory. We can obtain the corresponding $\mathcal R$ of a given $\mathcal T$ using map-matching methods, such as Markov-based methods~\cite{yang2018fast}.
\end{definition}

\subsection{Problem Statement}

\noindent \textbf{Pre-training Trajectory Embeddings.} 
Given a set of spatio-temporal trajectories, we aim to pre-train a trajectory encoder $f$ to generate an embedding vector $\boldsymbol e_{\mathcal T}\in \mathbb R^d$ when given a trajectory $\mathcal T$, i.e., $\boldsymbol e_{\mathcal T}=f(\mathcal T)$, where $d$ is the embedding dimensionality. 
\alert{
Our focus is on self-supervised pre-training of the encoder. The goal is to achieve embeddings that possess adaptability across a variety of downstream tasks.
Specifically, the learned embeddings aim to enhance the convergence speed and accuracy in diverse trajectory mining tasks, rendering them effective tools in an array of applications.
}

\subsection{Maximum Entropy Coding} \label{sec:maximum-entropy-coding}
Self-supervised pretext tasks are crucial for pre-training embedding methods. To ensure that embeddings generalize well across diverse downstream tasks, it is essential to minimize biases introduced during pre-training. 
The information theoretical principle of maximum entropy can be used for measuring the generality of embeddings.
According to this principle, the probability distribution that best represents the current state of knowledge about a system is the one with the largest entropy, in the context of precisely stated prior data~\cite{max-entropy-pri}. 
Leveraging this principle, maximum entropy coding (MEC)~\cite{liu2022self} aims to maximize the entropy of embeddings under a specific prior, \alert{with the end goal of producing embeddings that enhance performance across diverse tasks.}

Suppose we have a set of embeddings $\boldsymbol E\in \mathbb R^{d\times N}$, where $d$ is the embedding dimensionality and $N$ is the number of vectors.
To provide a tractable formulation of the entropy of the embeddings, MEC proposes the following coding length function:
\begin{equation}
    L=\frac{N+d}{2} \log \det (\boldsymbol I_N+\frac{d}{N\epsilon^2}\boldsymbol E^\top \boldsymbol E)
    \label{eq:coding-length-function}
\end{equation}
Here, $\boldsymbol I_N$ is an identity matrix with dimension $N$, and $\epsilon$ is the upper bound of the decoding error. This function calculates the coding length of a lossy data coding, which can be seen as an estimate of the entropy of the embeddings $\boldsymbol E$. The training objective under the MEC framework is to maximize Equation~\ref{eq:coding-length-function}.

\section{Methodology}\label{sec:methodology}

\begin{figure*}[t]
	\centering
	\includegraphics[width=.95\linewidth]{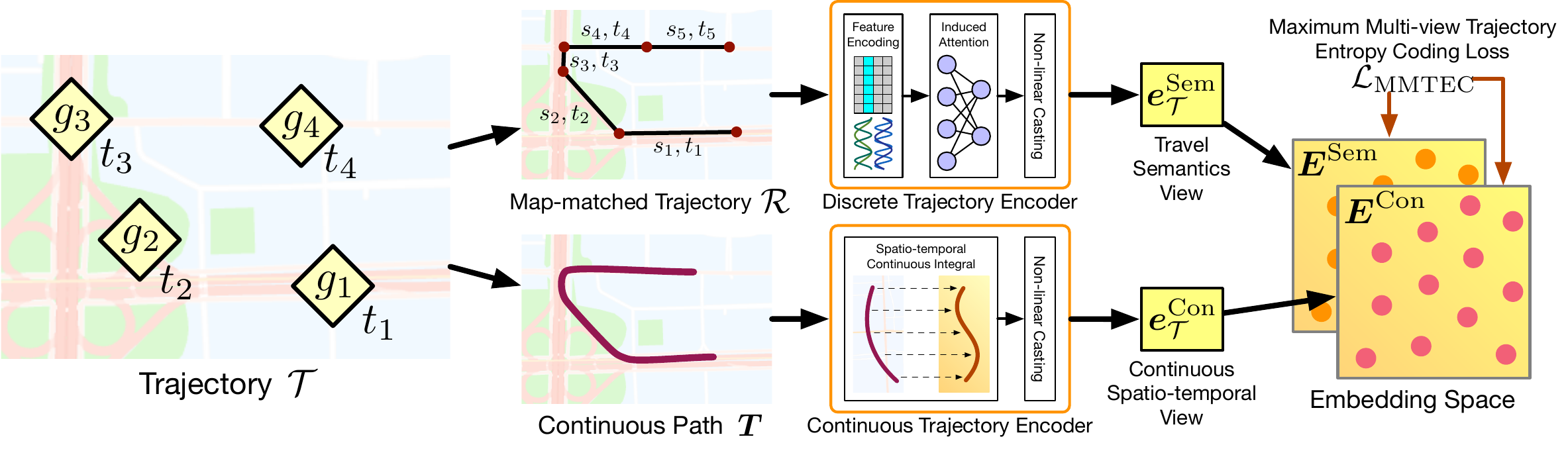}
	\caption{The overall framework of our proposed model.}
	\label{fig:framework}
\end{figure*}

\subsection{Overview}
Figure~\ref{fig:framework} shows the proposed framework. 
Given a spatio-temporal trajectory, our goal is to develop a self-supervised pretext task using maximum entropy coding, with the aim of maximizing the information entropy of the trajectory's embeddings. 
To incorporate multiple aspects of information embedded within the trajectory, we propose a multi-view scheme for modeling and fusing two aspects of information.

Specifically, we map-match the trajectory and use an attention-based discrete trajectory encoder to efficiently extract correlations from the map-matched trajectory, mapping it into the semantic embedding.
We also recover the underlying continuous movements of the trajectory and use a NeuralCDE-based continuous trajectory encoder to model the recovered movements and map these into the continuous spatio-temporal embedding. 
We incorporate these two embeddings into our pretext task by defining a view consistency prior constraint, which we use to pre-train the two embeddings with our final pretext task. After pre-training, we combine the two embeddings for use in downstream tasks. 
The remainder of the section provides a detailed explanation of the proposed method.

\subsection{Pre-training General Embeddings by Maximizing Entropy}
Current pre-training methods for trajectory embedding utilize generative or contrastive losses in their pretext tasks, which can introduce biases and restrict the generality of the learned embeddings to apply across different types of downstream tasks. 
To address this limitation, we draw inspiration from maximum entropy coding (MEC) as discussed in Section~\ref{sec:maximum-entropy-coding} and develop a pretext task that minimizes the bias introduced into pre-trained embeddings.

Suppose we have a set of trajectories $\{\mathcal T_1,\mathcal T_2, \dots, \mathcal T_N\}$ and the corresponding embeddings $\boldsymbol E_\mathcal T=\{\boldsymbol e_{\mathcal T_1}, \boldsymbol e_{\mathcal T_2},\dots, \boldsymbol e_{\mathcal T_N}\}$, where $\boldsymbol e_{\mathcal T_i}\in \mathbb R^d$. 
To minimize bias in the learned embeddings, we utilize MEC to maximize the entropy of the embeddings. Therefore, the negative of Equation~\ref{eq:coding-length-function} is used as our pre-training loss function. We simplify Equation~\ref{eq:coding-length-function} using Taylor expansion following~\cite{liu2022self} to obtain a more tractable loss function:
\begin{equation}
    \mathcal L=-\mathrm{trace}(\frac{N+d}{2} \sum_{k=1}^{\infty}\frac{(-1)^{k+1}}{k} (\frac{d}{N\epsilon^2} \boldsymbol E_\mathcal T^\top \boldsymbol E_\mathcal T)^k)
    \label{eq:expand-coding-length-function}
\end{equation}
According to the principle of maximum entropy, the resulting embeddings $\boldsymbol E_\mathcal T$ will have the maximum entropy among all possible embeddings. \alert{This characteristic facilitates optimal accuracy and convergence performance across diverse downstream tasks.}

Recall that the principle of maximum entropy requires a precisely stated prior. Simply pre-training trajectory embeddings with Equation~\ref{eq:expand-coding-length-function} without a well-defined prior will lead to all trajectories being represented uniformly~\cite{max-entropy-pri}. 
Additionally, training only one form of trajectory embedding limits the scope of their applicability. For example, embedding only the map-matched trajectories captures only the semantic information of a road network, disregarding the continuous spatio-temporal correlations of trajectories. A narrow focus can limit the performance of the embeddings on downstream tasks.

Based on these considerations, we propose a multi-view trajectory scheme that models different aspects of the information in trajectories. This allows us to introduce a testable prior based on the similarity between the different views, leading to more comprehensive and effective trajectory embeddings.

\subsection{Multi-view Trajectory Scheme}
We propose a multi-view trajectory scheme that aims to incorporates multiple aspects of trajectory information into embeddings. 
Specifically, we seek to incorporate both travel semantics and continuous spatio-temporal correlations. Instead of augmenting trajectories into pairs of samples, we model the two aspects separately using specifically designed trajectory encoders and construct intermediate trajectory embeddings.

\subsubsection{Embedding Travel Semantics}
\paratitle{Obtaining semantically rich representations using map-matching.}
To extract travel semantics from trajectories, we first map the trajectories onto the road network to utilize the semantic information in road segments. 
We use the Fast Map Matching (FMM) algorithm~\cite{yang2018fast} to obtain the corresponding map-matched trajectory $\mathcal R$ for a given trajectory $\mathcal T$, shown as:
\begin{equation}
    \mathcal R = \mathrm{FMM}(\mathcal T)
    \label{eq:fast-map-matching}
\end{equation}

The resulting $\mathcal R$ is a sequence of timestamped road segments, which is a semantically rich representation of the original $\mathcal T$. However, $\mathcal R$ is not appropriate as an embedding of $\mathcal T$ because it contains both discrete features (i.e., road segment indices) and numerical features (i.e., timestamps and road segment coordinates), and because different $\mathcal R$ have different lengths.

\begin{figure*}[t]
    \centering
    \includegraphics[width=1.0\linewidth]{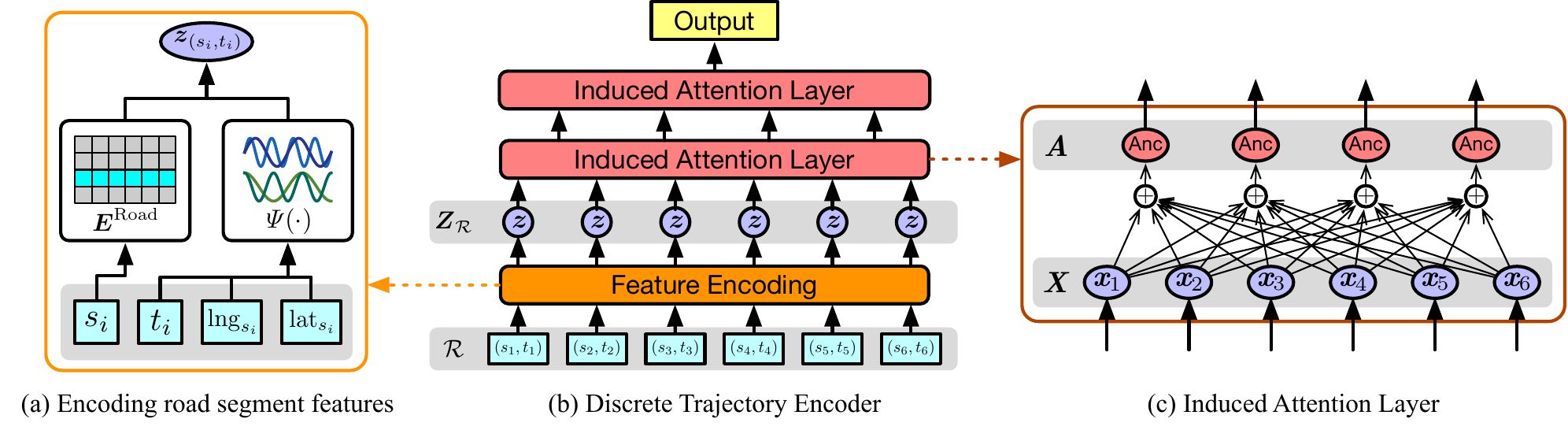}
    \caption{The architecture of our proposed discrete trajectory encoder and its components.}
    \label{fig:discrete-trajectory-encoder}
\end{figure*}

To address this, we propose an attention-based discrete encoder with a feature encoding layer and induced attention layers. This encoder efficiently and effectively models the correlations between road segments in $\mathcal R$ and maps $\mathcal R$ into a fixed-length embedding, as shown in Figure~\ref{fig:discrete-trajectory-encoder}. 
This encoder is realized as follows.

\paratitle{Encoding features in timestamped road segments.}
Each timestamped road segment in $\mathcal R$ contains several features: the road segment index $s_i$, the timestamp $t_i$ when the road segment is visited, and the spatial coordinates $(\mathrm{lng}_{s_i}, \mathrm{lat}_{s_i})$ of the road segment.

To embed the discrete road segment index $s_i$, we provide an index fetching embedding module so that the inherent semantic information in road segments can be captured and learned. Specifically, we initialize an embedding matrix $\boldsymbol E^{\mathrm{Road}}\in \mathbb R^{d\times |\mathbb S|}$, where each column $\boldsymbol E^{\mathrm{Road}}_{s_i}$ corresponds to the embedding vector of road segment $s_i$.

We then introduce a trigonometric function $\varPsi: \mathbb R \rightarrow \mathbb R^{d}$ to encode the numerical features $t_i, \mathrm{lng}_{s_i}, \mathrm{lat}_{s_i}$, thereby emphasizing their periodic characteristics, for example, the similarity in travel behavior during the same period of the day. 
The inspiration for this approach stems from both the positional encoding used in the Transformer~\cite{vaswani2017attention} and the temporal encoding proposed in recent studies~\cite{DBLP:conf/iclr/XuRKKA20,lin2021pre}. 
The encoding function $\varPsi(v)$ is defined as follows:
\begin{equation}
    \varPsi(v) = (\cos(\omega_1 v),\sin(\omega_1 v),\dots,\cos(\omega_{d/2} v),\sin(\omega_{d/2} v)),
    \label{eq:encode-numerical-features}
\end{equation}
where $v\in \{t_i\} \cup \{\mathrm{lng}_{s_i}\} \cup \{\mathrm{lat}_{s_i}\}, i\in\{1, 2, \dots, |\mathcal R|\}$ denotes the input numerical feature and $\omega_1,\omega_2,\dots,\omega_{d/2}$ are learnable parameters. 
Importantly, distinct sets of parameters are employed for different types of input features.
By using $\varPsi$, the encoder adeptly captures the spatial and temporal periodic characteristics of the numerical features via the $\sin$ and $\cos$ functions.  
\alert{Additionally, this encoding enhances the model's capture of relative correlations between features of different segments in $\mathcal R$. This is predominantly due to the central role played by dot-product computations in encoder architectures, including multi-head attention~\cite{vaswani2017attention}.} Specifically, the dot-product of two encoding vectors $\varPsi(v)$ and $\varPsi(v+\delta)$ is calculated as:
\begin{equation}
    \varPsi(v)\cdot\varPsi(v+\delta)
    = \cos(\omega_1\delta) + \cos(\omega_2\delta) + \dots + \cos(\omega_{d_v/2}\delta)
    \label{eq:dot-product-of-numerical-encoding}
\end{equation}
\alert{
This calculation illustrates that the distance (measured by the dot-product) between $\varPsi(v)$ and $\varPsi(v+\delta)$  relies exclusively on their numerical difference $\delta$, obviating the influence of offset $v$~\cite{DBLP:conf/nips/TancikSMFRSRBN20,DBLP:conf/nips/LiSLHB21}. 
Furthermore, it shows that the distance information is learnable, guided by the set of parameters $\{\omega_1,\dots,\omega_{d/2}\}$.
Our encoder then effectively captures this distance information, enabling the modeling of correlations between timestamped road segments within $\mathcal R$, such as the time gap between the traversal of two road segments.
}

Finally, we obtain the latent embedding of each timestamped road segment $(s_i, t_i)$, denoted as:
\begin{equation}
    \boldsymbol z_{(s_i, t_i)}=\boldsymbol E^{\mathrm{Road}}_{s_i} + \varPsi(t_i) + \varPsi(\mathrm{lng}_{s_i}) + \varPsi(\mathrm{lat}_{s_i})
\end{equation}

We can then transform a map-matched trajectory $\mathcal R$ into a sequence of latent embeddings:
\begin{equation}
    \boldsymbol Z_{\mathcal R}=\langle \boldsymbol z_{(s_1, t_1)},\boldsymbol z_{(s_2, t_2)}, \dots, \boldsymbol z_{(s_{|\mathcal R|}, t_{|\mathcal R|})} \rangle \in \mathbb R^{d\times |\mathcal R|}
\end{equation}

\paratitle{Efficient sequence encoder based on induced attention.}
To capture the correlation between road segments using a latent embedding sequence $\boldsymbol Z_{\mathcal R}$, an intuitive approach is to use the Transformer encoder, which has been adopted widely. 
However, the self-attention mechanism in the Transformer incurs quadratic time and memory complexity, meaning that it does not scale well to long sequences. 
Instead, we draw inspiration from recent developments in natural language processing for sequential models~\cite{lee2019set}. 
Specifically, we use an attention mechanism that incorporates long-term correlations and has low computational complexity. Our attention scheme, which we refer to as induced attention, utilizes globally-shared anchor sequences for attention weight calculation.

For the induced attention layer, we initialize an anchor sequence $\boldsymbol A_i\in\mathbb R^{d\times L_{A_i}}$ as the inherent parameter of this layer, where $i$ denotes the $i$-th layer and $L_{A_i}$ is the length of the anchor sequence. 
Then, the induced attention layer $\mathrm{IA}_i$ is realized using attention and feed-forward networks:
\begin{equation} 
\begin{aligned}
    \mathrm{IA}_i(\boldsymbol X_i) &= \mathrm{Norm}(\boldsymbol H_i + \mathrm{FFN}_i(\boldsymbol H_i)) \\
    \boldsymbol H_i &= \mathrm{Norm}(\boldsymbol A_i + \mathrm{Att}_i(\boldsymbol A_i, \boldsymbol X_i, \boldsymbol X_i)),
\end{aligned} 
\label{eq:induced-attention-layer}
\end{equation}
where $\mathrm{Norm}$, $\mathrm{FFN}_i$, and $\mathrm{Att}_i$ represent the layer normalization~\cite{ba2016layer}, the feed-forward network, and the multi-head dot-product attention of $i$-th layer, respectively. We use 8 attention heads. 
Next, $\boldsymbol X_i\in \mathbb R^{d\times L_{X_i}}$ is the input to $i$-th layer. During attention calculation, $\boldsymbol A_i$ is regarded as the query, while $\boldsymbol X_i$ is regarded as the key and value. 
The parameters in $\mathrm{FFN}_i, \mathrm{Att}_i$ and the anchor sequence $\boldsymbol A_i$ are shared for every specific $\mathrm{IA}$ layer between mini-batches, but are not shared between different layers.

We build our discrete encoder $\mathrm{DisEnc}$ by stacking two $\mathrm{IA}$ layers. Given a latent embedding sequence $\boldsymbol Z_{\mathcal R}$, the implementation of $\mathrm{DisEnc}$ is defined as:
\begin{equation}
    \mathrm{DisEnc}(\mathcal R) = \mathrm{IA}_2(
    \mathrm{IA}_1(\boldsymbol Z_{\mathcal R})),
\label{eq:discrete-encoder}
\end{equation}
where the anchor sequence length of the second $\mathrm{IA}$ layer $L_{A_2}$ is set to $1$ so that the output dimensionality of $\mathrm{DisEnc}$ is $d$.

By sharing the anchor sequences across mini-batches for attention score calculation rather than adopting self-attention, we build a more efficient discrete encoder that is suitable for modeling long-term correlations in map-matched trajectories.
\alert{
The computational complexity of Equation~\ref{eq:discrete-encoder} is $O(|\mathcal R|)$. In contrast, the complexity of a vanilla Transformer encoder is $O(|\mathcal R|^2)$, which does not scale well to large $|R|$.
An empirical study of the efficiency improvement achieved by the proposed model is covered in Section~\ref{sec:efficiency-comparison}.
}

\paratitle{Final semantic embedding.}
We obtain the final semantic embedding by appending a non-linear casting to the output of $\mathrm{DisEnc}$:
\begin{equation}
    \boldsymbol e_{\mathcal T}^\mathrm{Sem} = \boldsymbol W_2 (\sigma(\boldsymbol W_1\mathrm{DisEnc}(\mathcal R)),
    \label{eq:semantic-embedding-vector}
\end{equation}
where $\boldsymbol e_{\mathcal T}^\mathrm{Sem}\in\mathbb R^d$ denotes the final semantic embedding of trajectory $\mathcal T$, representing the travel semantics view. $\boldsymbol W_1, \boldsymbol W_2\in \mathbb R^{d\times d}$ are mapping matrices, and $\sigma$ denotes a non-linear activation function---we use ReLU. 

\subsubsection{Embedding Continuous Spatio-Temporal Correlations.}

\begin{figure}[t]
    \centering
    \includegraphics[width=0.9\linewidth]{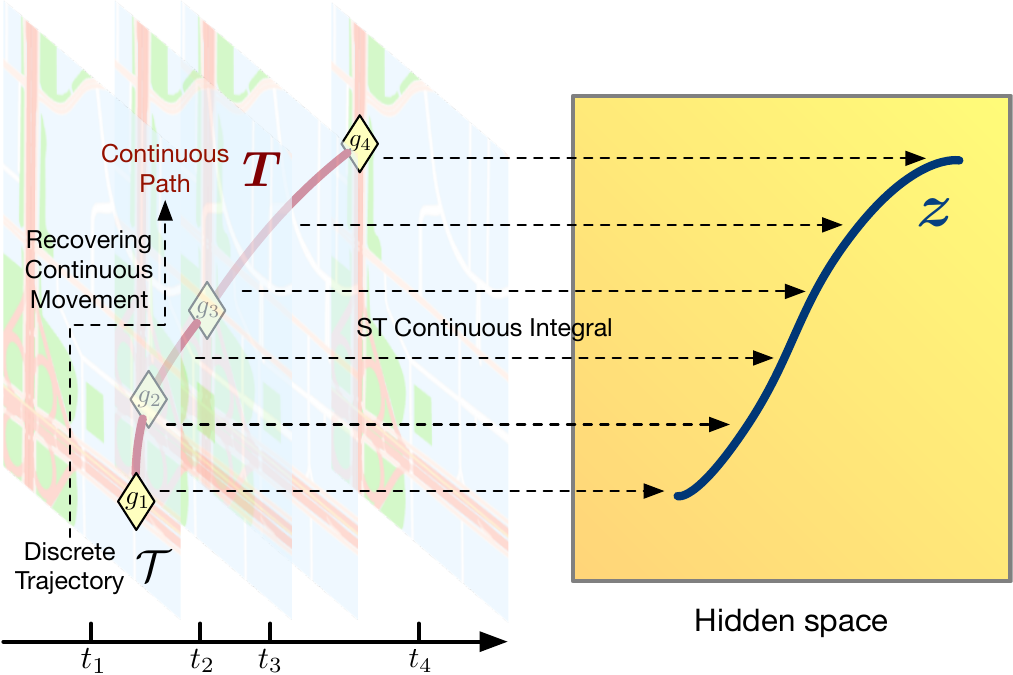}
    \caption{Illustration of the continuous trajectory encoder.}
    \label{fig:continuous-trajectory-encoder}
\end{figure}

\paratitle{Recovering continuous movements of trajectories.}
To properly model the spatio-temporal correlations in trajectories, it is essential to account for their continuous spatial and temporal dynamics. 
However, the irregular and discrete nature of GPS points in trajectories causes challenges for conventional sequential models such as RNNs~\cite{hochreiter1997long} and Transformers~\cite{vaswani2017attention}. 
Instead, we adopt an approach that can inherently consider the underlying continuous paths of trajectories, as shown in Figure~\ref{fig:continuous-trajectory-encoder}.

We first recover the continuous movements of trajectories based on their GPS points.
Given a trajectory $\mathcal T=\langle (g_1,t_1), (g_2,t_2), \dots, (g_{|\mathcal T|},t_{|\mathcal T|}) \rangle$, we approximate the underlying continuous movements of the trajectory by calculating its cubic spline:
\begin{equation}
    \boldsymbol T=\mathrm{Spline}(\mathcal T),
    \label{eq:natural-cubic-spline}
\end{equation}
where $\mathrm{Spline}$ is the cubic Hermite spline that interpolates the underlying continuous process of the trajectory. In particular, $\boldsymbol T$ is the natural cubic spline with knots at $t_1,t_2,\dots, t_{|\mathcal T|}$ that satisfies:
\begin{equation}
\begin{split}
    &\color{black}{\boldsymbol T: t\rightarrow \mathbb R^2,~t\in[t_1, t_{|\mathcal T|}],}\\
    &\boldsymbol T_{t_i}=g_i, ~i\in\{1,2,\dots,|\mathcal T|\}
\end{split}
\label{eq:natural-cubic-spline-mapping}
\end{equation}

Since $\boldsymbol{T}$ estimates the underlying continuous movement of a trajectory $\mathcal{T}$, it can serve as an intermediate step for mapping $\mathcal{T}$ to the continuous spatio-temporal embedding. However, $\boldsymbol{T}$ is unsuitable for most trajectory analysis tasks due to its continuous definition. 
To address this, we propose a continuous encoder $\mathrm{ConEnc}$ that can capture the continuous dynamics represented by $\boldsymbol{T}$.

\paratitle{Spatio-temporal continuous integral.} 
To construct the continuous encoder $\mathrm{ConEnc}$, we leverage the family of ordinary differential equation methods, specifically the neural controlled differential equation (NeuralCDE)~\cite{kidger2020neural}. This method captures the underlying sequential processes in continuous time and models both the input features and hidden states in a continuous manner. 
We adopt an existing formulation~\cite{kidger2020neural} and define our continuous encoder as follows:
\begin{equation}
\begin{split}
    \mathrm{ConEnc}(\boldsymbol T)
    &=\boldsymbol z_{t_1} + \int_{t_1}^{t_{|\mathcal T|}}\mathrm{CDENet}(\boldsymbol z_t)\mathrm d \boldsymbol T_t \\
    &=\boldsymbol z_{t_1} + \int_{t_1}^{t_{|\mathcal T|}} \mathrm{CDENet}(\boldsymbol z_t)\frac{\mathrm d \boldsymbol T}{\mathrm d t}(t)\mathrm d t
\end{split}
\label{eq:neural-cde}
\end{equation}
\alert{
Here, $\frac{\mathrm d \boldsymbol T}{\mathrm d t}$ denotes the derivative of the cubic spline with regard to time, which we implement by employing the \texttt{torchdiffeq} package~\cite{torchdiffeq}.
}
Next, $\boldsymbol z_{t_1}=\mathrm{InitNet}(g_1,t_1)$ is a parameterized initial state, $\mathrm{InitNet}:\mathbb R^3\rightarrow \mathbb R^d$ denotes the neural network used for calculating the initial state, and $\mathrm{CDENet}: \mathbb R^d\rightarrow \mathbb R^{d\times 3}$ is a neural network that serves as the integral kernel of CDE. 
$\mathrm{InitNet}$ and $\mathrm{CDENet}$ are realized as follows.

\paratitle{Parameterized CDE initial state and integral kernel.}
The initial state $\boldsymbol z_{t_1}$ determines the starting point of the CDE integral and depends on the first timestamped GPS point $(g_1,t_1)$ of a trajectory.
Since all features in a GPS point are numerical, we calculate the initial state using the trigonometric function from Equation~\ref{eq:encode-numerical-features}. Thus, the neural network $\mathrm{InitNet}$ is defined as:
\begin{equation}
    \mathrm{InitNet}(g_1, t_1) = \varPsi(\mathrm{lng}_1) + \varPsi(\mathrm{lat}_1) + \varPsi(t_1)
\end{equation}

The neural network $\mathrm{CDENet}$ serves as the CDE integral kernel. Here, we use a two-layer non-linear mapping:
\begin{equation}
    \mathrm{CDENet}(\boldsymbol z_t) = \boldsymbol W_6 (\sigma(\boldsymbol W_5 \boldsymbol z_t + \boldsymbol b_5)) + \boldsymbol b_6,
\end{equation}
where $\boldsymbol W_5 \in \mathbb R^{3d\times d}$ and $\boldsymbol W_6\in \mathbb R^{3d \times 3d}$ are mapping matrices and $\boldsymbol b_5, \boldsymbol b_6 \in \mathbb R^{3d}$ are biases.

\paratitle{Final continuous spatio-temporal embedding.}
Akin to Equation~\ref{eq:semantic-embedding-vector}, we pair the continuous encoder with a non-linear casting to obtain the continuous spatio-temporal embedding vector:
\begin{equation}
    \boldsymbol e_{\mathcal T}^\mathrm{Con}=\boldsymbol W_4(\sigma(\boldsymbol W_3 \mathrm{ConEnc}(\boldsymbol T))),
    \label{eq:continuous-st-embedding-vector}
\end{equation}
where $\boldsymbol e_{\mathcal T}^\mathrm{Con}$ denotes the final continuous spatio-temporal embedding of trajectory $\mathcal T$, representing the continuous spatio-temporal view and $\boldsymbol W_3, \boldsymbol W_4\in \mathbb R^{d\times d}$ are mapping matrices. 

\subsubsection{Combining the Intermediate Embeddings}
The final embedding of a trajectory $\mathcal{T}$ is obtained by combining the semantic and continuous spatio-temporal embeddings as follows:
\begin{equation}
    \boldsymbol e_\mathcal T = \boldsymbol e_\mathcal T^\mathrm{Sem} \oplus \boldsymbol e_\mathcal T^\mathrm{Con},
    \label{eq:combine-intermediate-embeddings}
\end{equation}
where $\oplus$ denotes the concatenation operation along feature dimension. 
\alert{
The resulting embeddings incorporates both informational aspects captured in trajectories. This enables a comprehensive modeling of the informational aspects of trajectories, yielding embeddings that are more capable of enhancing the prediction performance of downstream tasks.
}

\subsection{Maximum Multi-view Trajectory Entropy Coding}
To ensure that we satisfy the premise of the principle of maximum entropy and learn meaningful trajectory embeddings, we introduce testable prior information to Equation~\ref{eq:expand-coding-length-function} based on the two embeddings obtained in Equations~\ref{eq:semantic-embedding-vector} and~\ref{eq:continuous-st-embedding-vector}.

Given a mini-batch of trajectories $\{\mathcal T_1,\mathcal T_2, \dots, \mathcal T_N\}$ used during batch gradient descent training, where $N$ is the batch size, 
we obtain their corresponding semantic embeddings $\boldsymbol E^{\mathrm{Sem}}=\{\boldsymbol e_{\mathcal T_1}^{\mathrm{Sem}}, \boldsymbol e_{\mathcal T_2}^{\mathrm{Sem}}, \dots, \boldsymbol e_{\mathcal T_N}^{\mathrm{Sem}}\} \in \mathbb R^{d\times N}$
and their corresponding continuous spatio-temporal embeddings 
$\boldsymbol E^{\mathrm{Con}}=\{\boldsymbol e_{\mathcal T_1}^{\mathrm{Con}}, \boldsymbol e_{\mathcal T_2}^{\mathrm{Con}}, \dots, \boldsymbol e_{\mathcal T_N}^{\mathrm{Con}}\} \in \mathbb R^{d\times N}$.
As the two types of embeddings correspond to the same set of trajectories, they can be regarded as different views on that set.
To incorporate the view consistency prior between $\boldsymbol E^{\mathrm{Sem}}$ and $\boldsymbol E^{\mathrm{Con}}$ into Equation~\ref{eq:expand-coding-length-function}, we obtain:
\begin{equation}
    \mathcal L'=-\mathrm{trace}(\frac{N+d}{2} \sum_{k=1}^{\infty}\frac{(-1)^{k+1}}{k} (\frac{d}{N\epsilon^2} {\boldsymbol E^{\mathrm{Sem}}}^\top \boldsymbol E^{\mathrm{Con}})^k)
    \label{eq:prior-coding-length-function}
\end{equation}

To make the calculation feasible, we approximate Equation~\ref{eq:prior-coding-length-function} by truncating the infinite Taylor expansion, obtaining the following MMTEC loss function:
\begin{equation}
    \mathcal L_\mathrm{MMTEC}=-\mathrm{trace}(\frac{N+d}{2} \sum_{k=1}^{K}\frac{(-1)^{k+1}}{k} (\frac{d}{N\epsilon^2} {\boldsymbol E^{\mathrm{Sem}}}^\top \boldsymbol E^{\mathrm{Con}})^k)
    \label{eq:mmtec-loss-function}
\end{equation}

Here, $K$ is the order of the Taylor expansion, which is a hyper-parameter. 
This loss function maximizes the entropy of embeddings while ensuring view consistency. 
\alert{
Complemented by the incorporation of both aspects of trajectory information, as captured in  Equation~\ref{eq:combine-intermediate-embeddings}, the learned embeddings are able to enhance the convergence speed and accuracy of a range of downstream tasks.
}

\section{Experiments}\label{sec:experiments}
To evaluate proposed method for learning trajectory embeddings, we conduct experiments on two real-world datasets and apply the learned embeddings in three downstream tasks. We compare our results with those obtained using other state-of-the-art methods.

\subsection{Datasets}
We conduct the experiments on two real-world datasets, Chengdu and Xian, that contain taxi trip trajectories collected from the central areas of the respective cities. The datasets are provided by DiDi\footnote{The dataset was published at https://gaia.didichuxing.com/.}. 
To ensure a reasonable size of the dataset, we sort the drivers descendingly according to how many trajectories they are associated with. We then select the trajectories produced by the top 20,000 drivers. 
We also download the road networks of Chengdu and Xian from OpenStreetMap (OSM)~\cite{osm} for use in the map matching presented in Equation~\ref{eq:fast-map-matching}. We retain only the road segments covered by at least one trajectory. Table~\ref{table:dataset-statistics} presents statistics of the datasets.

\begin{table}[h]
	\centering
	\caption{Statistics of datasets.}
	\scalebox{1}{
		\begin{tabular}{cccc}
			\toprule
			Dataset & \#Trajectories & \#Segments & \#Records \\
			\midrule
                Chengdu & 298,995 & 3,791 & 7,025,468 \\
                Xian & 376,407 & 3,558 & 10,198,837 \\
			\bottomrule
		\end{tabular}
	}
	\label{table:dataset-statistics}
\end{table}

\subsection{Comparison Methods}
To assess the benefits of the proposed model over the state-of-the-art, we include comparisons with six state-of-the-art pre-training trajectory embedding methods.

\begin{itemize}
\item \textbf{traj2vec}~\cite{yao2017trajectory}: constructs feature sequences by calculating the differences in the spatio-temporal attributes between consecutive trajectory points and applies an auto-regressive pretext task to learn embeddings.
\item \textbf{t2vec}~\cite{li2018deep}: pre-trains an RNN trajectory encoder based on a de-noising auto-encoder, aiming to recover fine-grained trajectories given sparse ones.
\item \textbf{GM-VSAE}~\cite{liu2020online}: casts trajectories into a multi-dimension Gaussian space using an RNN variational trajectory encoder and pre-trains the model following variational auto-encoders.
\item \textbf{TremBR}~\cite{fu2020trembr}: constructs an RNN auto-encoder to learn trajectory embeddings with both road segment and timestamp recovery tasks.
\item \textbf{CTLE}~\cite{lin2021pre}: pre-trains a bi-directional Transformer with two MLM-based tasks and then calculates the embeddings of locations based on their contexts. We apply a mean pooling on the embeddings of all points in a trajectory to get its embedding.
\item \alert{\textbf{PT2Vec}~\cite{li2023trajectory}: combines a road network partitioning and an RNN auto-encoder to effectively learn trajectory embeddings on large-scale road networks.}
\item \textbf{Toast}~\cite{chen2021robust}: applies the node2vec model to the road network to pre-train a set of embeddings for the road segments and then pre-trains a Transformer encoder with an MLM-based task and a sequence discrimination task to generate trajectory embeddings.
\end{itemize}

\alert{
Table~\ref{table:related-works-comparison} offers a comparison of three fundamental attributes of the pre-training trajectory embedding methods considered in the experiments.
In the table, the column labeled \textit{Initial task} refers to the downstream tasks the methods targeted in within their respective publications, while the column titled \textit{Pre-training framework} denotes the pre-training approach embraced by each method.
}

\begin{table}[h]
\centering
\caption{\alert{Comparison of pre-training trajectory embedding methods.}}
\scalebox{.87}{
    \begin{tabular}{c|ccc}
        \toprule
        \multirow{2}{*}{\alert{Methods}} & \alert{Encoder} & \multirow{2}{*}{\alert{Initial task}} & \alert{Pre-training} \\
        & \alert{structure} & & \alert{framework} \\
        \midrule
        \alert{traj2vec} & \alert{RNN} & \alert{Clustering} & \alert{Auto-encoding} \\
        \alert{t2vec} & \alert{RNN} & \alert{Similarity Measurement} & \alert{Auto-encoding} \\
        \alert{GM-VSAE} & \alert{RNN} & \alert{Anomaly Detection} & \alert{Variational AE} \\
        \alert{TremBR} & \alert{RNN} & \alert{Embedding Learning} & \alert{Auto-encoding} \\
        \alert{CTLE} & \alert{Transformer} & \alert{Destination Prediction} & \alert{MLM} \\
        \alert{PT2Vec} & \alert{RNN} & \alert{Similarity Measurement} & \alert{Auto-encoding} \\
        \alert{Toast} & \alert{Transformer} & \alert{Embedding Learning} & \alert{MLM+Contrastive} \\
        \alert{MMTEC} & \alert{Multi-view} & \alert{Embedding Learning} & \alert{Max Entropy} \\
        \bottomrule
        \end{tabular}
}
\label{table:related-works-comparison}
\end{table}

\subsection{Downstream Tasks}
To assess the generality of the trajectory embeddings learned by our method, we evaluate their performance on three downstream tasks: similar trajectory search, travel time estimation, and destination prediction.

\paratitle{Similar trajectory search.} 
Our goal is to rank similar trajectories based on their distances in the learned embedding space. Since ground truth labels are not available, we adopt the evaluation strategy from Toast~\cite{chen2021robust} and randomly select 10,000 trajectories from the full dataset as the target trajectories, denoted as $\mathbb T$. 
For each $\mathcal T_i \in \mathbb T$, we randomly select a 20\% portion, and replace it with a shortest path on the road network from the origin to the destination of the portion, ensuring it is distinct from the original portion, this way obtaining a positive sample $\mathcal T_i'$.
We then calculate the embedding $\boldsymbol e_{\mathcal T_i'}$ of $\mathcal T_i'$ and determine its similarity with all target trajectories $\mathcal T_j\in \mathbb T$ in the embedding space using cosine similarity. The target trajectory most similar $\mathcal T_i'$, i.e., the ground truth, should be $\mathcal T_i$.

\paratitle{Travel time estimation.}
Our aim is to estimate the travel time of trajectories using their embeddings. We feed the embedding methods with trajectories excluding timestamps and then pass the resulting embeddings to a fully-connected network for travel time regression.

\paratitle{Destination prediction.}
Our aim is to predict the destination road segments for trajectories using their embeddings.
We feed the embedding methods a map-matched trajectory $\mathcal R$ excluding its last 5 segments. 
The output embedding vector is then passed to a fully-connected network to classify the last road segment index of $\mathcal R$.

\begin{table}[h]
    \centering
    \caption{Hyper-parameter ranges and optimal values.}
    \begin{threeparttable}
        \begin{tabular}{c|c}
            \toprule
            Parameter & Range \\
            \midrule
            $d$ & 32, \underline{64}, 128, 192, 256 \\
            $N$ & 128, 256, 384, \underline{512}, 640, 768, 896, 1024 \\
            $\epsilon^2$ & 128, 256, 384, \underline{512}, 640, 768, 896, 1024 \\
            $L_{A_1}$ & 0, 4, \underline{8}, 12, 16, 20, 24, 28, 32\\
            $K$ & 1, 2, 3, 4, \underline{5}, 6, 7, 8, 9, 10\\
            \bottomrule
        \end{tabular}
        \begin{tablenotes}\footnotesize
            \item[]{\underline{Underline} denotes the optimal value.}
        \end{tablenotes}
    \end{threeparttable}
    \label{table:hyper-parameters}
\end{table}

\begin{table*}[t]
\centering
\caption{Performance comparison of trajectory embedding methods on the Chengdu dataset.}
\begin{threeparttable}
    \begin{tabular}{c|cc|ccc|ccc}
    \toprule
    Task & \multicolumn{2}{c|}{Similar Trajectory Search} & \multicolumn{3}{c|}{Travel Time Estimation} & \multicolumn{3}{c}{Destination Prediction} \\
    \midrule
    Metric & Acc@1(\%) & Acc@5(\%) & MAE(min) &
        RMSE(min) & MAPE(\%) & Acc@1(\%) & 
        Acc@5(\%) & F1(\%) \\
    \midrule
    traj2vec & 61.38$\pm$0.78 & 71.19$\pm$0.59 & 
        2.570$\pm$0.022 & 3.740$\pm$0.020 & 35.11$\pm$0.24 & 
        45.54$\pm$0.27 & 68.23$\pm$0.20 & 17.83$\pm$0.09 \\
    t2vec & 61.52$\pm$1.14 & 71.70$\pm$1.56 & 
        2.518$\pm$0.024 & 3.695$\pm$0.031 & 33.49$\pm$0.30 & 
        39.52$\pm$0.13 & 63.36$\pm$0.40 & 13.51$\pm$0.12 \\
    GM-VSAE & 64.28$\pm$0.87 & 76.17$\pm$0.49 & 
        2.515$\pm$0.033 & 3.732$\pm$0.041 & 33.49$\pm$0.32 & 
        44.37$\pm$0.63 & 67.30$\pm$0.86 & 17.64$\pm$0.11 \\
    TremBR & 71.51$\pm$2.51 & 79.49$\pm$1.65 & 
        2.445$\pm$0.043 & 3.633$\pm$0.063 & 32.01$\pm$0.45 & 
        55.04$\pm$0.55 & 77.74$\pm$0.26 & 26.89$\pm$0.13 \\
    CTLE & 73.75$\pm$1.41 & 82.69$\pm$1.62 & 
        2.341$\pm$0.035 & 3.558$\pm$0.071 & 30.37$\pm$0.51 & 
        59.88$\pm$0.26 & 79.92$\pm$0.13 & 27.66$\pm$0.13 \\
    \alert{PT2Vec} & \alert{75.27$\pm$1.18} & \alert{85.03$\pm$1.48} & 
        \alert{2.391$\pm$0.027} & \alert{3.618$\pm$0.039} & \alert{30.64$\pm$0.36} & 
        \alert{54.98$\pm$0.91} & \alert{75.27$\pm$0.82} & \alert{24.19$\pm$0.42} \\
    Toast & 78.67$\pm$0.71 & 86.40$\pm$0.60 & 
        2.320$\pm$0.028 & 3.477$\pm$0.069 & 29.44$\pm$0.44 & 
        57.99$\pm$0.64 & 77.14$\pm$0.85 & 26.17$\pm$0.99 \\
    \midrule
    generative & 72.89$\pm$0.84 & 83.20$\pm$1.21 & 
        \underline{2.239$\pm$0.042} & \underline{3.341$\pm$0.033} & \underline{28.10$\pm$0.93} & 
        \underline{62.52$\pm$0.12} & \underline{89.26$\pm$0.14} & \underline{30.04$\pm$0.14} \\
    contrastive & \underline{80.62$\pm$1.99} & \underline{93.13$\pm$1.41} & 
        2.419$\pm$0.124 & 3.590$\pm$0.184 & 31.59$\pm$1.79 & 
        60.84$\pm$0.31 & 87.66$\pm$0.13 & 27.35$\pm$0.28 \\
    \textbf{MMTEC} & \textbf{84.27$\pm$3.02} & \textbf{95.15$\pm$1.63} & 
        \textbf{2.079$\pm$0.109} & \textbf{3.104$\pm$0.194} & \textbf{26.02$\pm$1.00} & 
        \textbf{64.61$\pm$0.13} & \textbf{91.71$\pm$0.06} & \textbf{33.35$\pm$0.11} \\
    \bottomrule
    \end{tabular}
    \begin{tablenotes}\footnotesize
        \item[]{\textbf{Bold} denotes the best result, 
        \underline{underline} denotes the second best result.}
    \end{tablenotes}
\end{threeparttable}
\label{table:chengdu-overall-result}
\end{table*}

\begin{table*}[t]
\centering
\caption{Performance comparison of trajectory embedding methods on the Xian dataset.}
\begin{threeparttable}
    \begin{tabular}{c|cc|ccc|ccc}
    \toprule
    Task & \multicolumn{2}{c|}{Similar Trajectory Search} & \multicolumn{3}{c|}{Travel Time Estimation} & \multicolumn{3}{c}{Destination Prediction} \\
    \midrule
    Metric & Acc@1(\%) & Acc@5(\%) & 
        MAE(min) & RMSE(min) & MAPE(\%) & 
        Acc@1(\%) & Acc@5(\%) & F1(\%) \\
    \midrule
    traj2vec & 58.57$\pm$0.72 & 77.03$\pm$0.62 & 
        3.142$\pm$0.050 & 5.159$\pm$0.048 & 30.64$\pm$0.79 & 
        44.04$\pm$0.43 & 67.20$\pm$0.21 & 13.56$\pm$0.49 \\
    t2vec & 58.19$\pm$1.27 & 73.78$\pm$1.14 & 
        3.178$\pm$0.016 & 5.248$\pm$0.011 & 30.93$\pm$0.30 & 
        37.99$\pm$0.16 & 62.47$\pm$0.41 & 8.73$\pm$0.34 \\
    GM-VSAE & 62.47$\pm$0.30 & 78.24$\pm$1.03 & 
        3.143$\pm$0.032 & 5.123$\pm$0.039 & 30.73$\pm$0.43 & 
        45.21$\pm$0.89 & 68.82$\pm$0.88 & 13.73$\pm$0.79 \\
    TremBR & 65.55$\pm$1.34 & 79.55$\pm$1.63 & 
        3.037$\pm$0.028 & 5.073$\pm$0.022 & 28.65$\pm$0.31 & 
        56.78$\pm$0.61 & 79.33$\pm$0.98 & 22.14$\pm$0.90 \\
    CTLE & 65.74$\pm$1.95 & 79.26$\pm$1.25 & 
        2.949$\pm$0.042 & 4.912$\pm$0.108 & 27.74$\pm$0.58 & 
        62.25$\pm$0.53 & 85.21$\pm$0.53 & 27.17$\pm$0.81 \\
    \alert{PT2Vec} & \alert{68.64$\pm$1.40} & \alert{80.34$\pm$1.79} & 
        \alert{3.000$\pm$0.027} & \alert{4.972$\pm$0.062} & \alert{28.11$\pm$0.53} & 
        \alert{56.88$\pm$1.28} & \alert{79.14$\pm$1.40} & \alert{22.61$\pm$1.27} \\
    Toast & 69.43$\pm$2.12 & 83.41$\pm$1.67 & 
        2.941$\pm$0.066 & 4.894$\pm$0.173 & \underline{26.86$\pm$0.80} & 
        59.47$\pm$0.65 & 81.77$\pm$0.71 & 24.71$\pm$0.93 \\
    \midrule
    generative & 66.95$\pm$1.93 & 82.28$\pm$1.24 & 
        \underline{2.900$\pm$0.090} & \underline{4.724$\pm$0.138} & 26.91$\pm$0.92 & 
        \underline{64.52$\pm$0.18} & \underline{92.09$\pm$0.11} & \underline{30.14$\pm$0.74} \\
    contrastive & \textbf{77.27$\pm$1.65} & \underline{89.92$\pm$1.22} & 
        3.158$\pm$0.049 & 5.086$\pm$0.111 & 29.47$\pm$1.38 & 
        62.50$\pm$0.14 & 87.46$\pm$0.07 & 26.75$\pm$0.41 \\
    \textbf{MMTEC} & \underline{76.57$\pm$3.05} & \textbf{91.38$\pm$2.04} & 
        \textbf{2.870$\pm$0.023} & \textbf{4.653$\pm$0.073} & \textbf{26.06$\pm$0.44} & 
        \textbf{67.62$\pm$0.21} & \textbf{93.68$\pm$0.08} & \textbf{36.38$\pm$0.79} \\
    \bottomrule
    \end{tabular}
    \begin{tablenotes}\footnotesize
        \item[]{\textbf{Bold} denotes the best result, 
        \underline{underline} denotes the second best result.}
    \end{tablenotes}
\end{threeparttable}
\label{table:xian-overall-result}
\end{table*}

\subsection{Settings}
For all datasets, we sort trajectories by their start time and split the whole trajectory set into training, evaluation, and testing sets by 8:1:1. 
Both pre-training embedding methods and downstream predictors are trained with the training set. The embedding methods are pre-trained for 30 epochs, while the downstream predictors are stopped early on the evaluation set. The final metrics are calculated on the testing set.
We use Top-$N$ accuracy (i.e., Acc@$N$, $N=1, 5$) for the similar trajectory search task; Acc@$N$, $N=1, 5$, and macro-F1 for the destination prediction task; and mean absolute error (MAE), root mean square error (RMSE), and mean absolute percentage error (MAPE) for the travel time estimation task.
We run each set of experiments 10 times and report the mean and deviation of the metrics.

All models are implemented using PyTorch~\cite{paszke2019pytorch}.\footnote{The code and sample datasets are available in the supplementary material, and will be published on GitHub after the review.}
The proposed method has five key hyper-parameters, whose ranges and optimal values are listed in Table~\ref{table:hyper-parameters}. We choose parameters based on the Acc@1 results of the similar trajectory search task on the validation set of the Chengdu dataset. We report the effectiveness of these parameters in Section~\ref{sec:impact-of-hyper-parameters}.

\subsection{Experimental Results}
\subsubsection{Overall Performance}

Tables~\ref{table:chengdu-overall-result} and \ref{table:xian-overall-result} present a comprehensive comparison of the performance of different trajectory embedding methods across the three tasks. 
We also provide a visual illustration of the results in Figure~\ref{fig:radar}. Our proposed method consistently outperforms the other methods, offering evidence that it is the best at generating general and comprehensive trajectory embeddings.

\paratitle{Generality of the pretext task.}
traj2vec, t2vec, GM-VSAE, TremBR, \alert{and PT2Vec} all adopt the auto-encoding or auto-regressive frameworks. 
Among these, t2vec re-samples the original trajectories and adds noise to the encoder input to make the learned embeddings more robust to sparse and noisy trajectories. GM-VSAE uses a multi-dimensional Gaussian space as its latent embedding space and implements a VAE-like encoder-decoder architecture~\cite{kingma2013auto} to better model the trajectory distribution. 
However, auto-regressive pre-training methods focus mainly on recovering the detailed raw features of trajectories, biasing the learned embeddings towards point-level downstream tasks, such as travel time estimation and destination prediction. As the results show, auto-regressive pre-trained embedding methods perform poorly on sequence-level tasks, like similar trajectory search.

CTLE and Toast use the masked language model (MLM) pretext task from BERT~\cite{devlin2018bert}. 
Compared to the auto-regressive task, MLM utilizes the bi-directionality of Transformers and helps the model better understand the correlations between points in a trajectory. 
However, as a generative pretext task, it also biases the learned embeddings towards certain types of downstream tasks. 
Interestingly, Toast additionally implements a sequence-level pretext task
that helps Toast perform better on the similar trajectory search task.

The pretext task of our proposed method is built based on the principle of maximum entropy, aiming to learn embeddings that generalize well across different types of downstream tasks. This makes our method better suited for pre-training general trajectory embeddings. We further investigate the effectiveness of pretext tasks by pre-training our embedding models generatively and contrastively---the detailed analysis is presented in Section~\ref{sec:effectiveness-of-the-pretext-task}.

\paratitle{Comprehensiveness of the incorporated information.}
Methods traj2vec, t2vec, and GM-VSAE are designed to model the spatio-temporal features of trajectories based on GPS coordinates and timestamps, but they do not consider the semantic information in the underlying road networks. This limitation reduces their performance on tasks such as destination prediction, which requires predicting destination road segments. Additionally, their trajectory encoders are based on RNNs, which are unable to effectively capture continuous spatio-temporal correlations.

In comparison, TremBR, CTLE, \alert{PT2Vec}, and Toast incorporate semantic information by mapping trajectories onto road networks and using trajectory encoders that encode map-matched trajectories into embeddings. TremBR also considers temporal information by including the visiting time of road segments as an input feature. CTLE incorporates relative and absolute temporal information through a temporal encoding layer and a masked time pre-training task. \alert{PT2Vec applies a graph partition algorithm to road networks to build a more effective model on large-scale networks.} Toast improves its incorporation of semantic information by pre-training embeddings for the road network using node2vec. However, their trajectory encoders still rely on RNNs or Transformers, which update their hidden states discretely, making it challenging to accurately model the continuous spatio-temporal correlations in trajectories.

In contrast, our method incorporates both the semantic information of travel and the continuous spatio-temporal correlations present in trajectories. This enables comprehensive information to be incorporated into the learned embeddings, thereby improving the performance of downstream tasks.

\paratitle{Performance comparison summary.}
The proposed method leverages the benefits of maximum entropy coding and the trajectory multi-view  approach to form the MMTEC pretext task, which enables the learning of trajectory embeddings that are highly generalizable across different downstream tasks. 
The proposed method captures both the travel semantic information and continuous spatio-temporal correlations in trajectories, providing comprehensive information that improves the performance of downstream tasks. 
Furthermore, the method's discrete and continuous trajectory encoders effectively and efficiently embed trajectories. These advantages yield superior performance across multiple downstream tasks.

\subsubsection{Efficiency Comparison} \label{sec:efficiency-comparison}

\begin{table}[t]
    \centering
    \caption{Efficiency comparison between methods on the Chengdu dataset.}
    \label{table:efficiency}
    \begin{threeparttable}
    \begin{tabular}{c|ccc}
        \toprule
        \multirow{2}{*}{Property} & Model size & Train speed & Embed time \\
        & (MBytes) & (min/epoch) & (sec) \\
        \midrule
        traj2vec & \underline{\textbf{1.106}} & 0.924 & 1.039 \\
        t2vec & \underline{\textbf{1.106}} & 0.851 & \underline{1.028} \\
        GM-VSAE & 3.029 & 0.931 & 1.482 \\
        TremBR & 6.711 & 0.973 & 1.428 \\
        CTLE & 3.218 & 0.795 & 1.132 \\
        \alert{PT2Vec} & \alert{2.462} & \alert{0.874} & \alert{1.231} \\
        Toast & 3.218 & 2.733 & 1.129 \\
        \alert{Transformer} & \alert{1.909} & \alert{\underline{0.713}} & \alert{1.033} \\
        \textbf{MMTEC} & 1.761 & \textbf{0.371} & \textbf{0.510} \\
        \bottomrule
    \end{tabular}
    \begin{tablenotes}\footnotesize
        \item[]{\textbf{Bold} denotes the best result, 
        \underline{underline} denotes the second best result.}
    \end{tablenotes}
    \end{threeparttable}
\end{table}

When investigating the efficiency of the trajectory embedding methods, we consider the model size, training speed, and embedding time of all embedding methods on the Chengdu dataset. 
The training speed is the time it takes to complete one epoch of training, and the embedding time is the time in seconds it takes to calculate embeddings for all trajectories in the test set. The results are presented in Table~\ref{table:efficiency}.

Among the methods that incorporate road networks, our method has the smallest model size. This is because we use maximum entropy coding and multi-view consistency to reduce redundancy in the learned trajectory embeddings.

The RNN-based methods, such as traj2vec, t2vec, GM-VSAE, TremBR, \alert{and PT2Vec}, are relatively slower at training and embedding, as RNNs do not parallelize on long trajectory sequences. In comparison, the Transformer-based embedding method CTLE is faster at training thanks to its parallelized attention mechanism. Toast is the slowest at training as the pre-training of road segment embeddings takes up additional time. 
Our method is the fastest at training and embedding due to its sharing of anchor sequences across mini-batches.
\alert{
To assess the efficiency of the proposed discrete trajectory encoder, we replace the induced attention layers within the encoder by vanilla Transformer encoder layers. 
The efficiency comparison of the modified encoder is shown in the \textit{Transformer} row in Table~\ref{table:efficiency}.
A comparison of the efficiency metrics with those of MMTEC indicates that a robust acceleration is achieved by the incorporation of the induced attention layers.
}

Overall, our method is efficient while outperforming the other trajectory embedding methods on downstream tasks.

\begin{figure}[t]
  \centering
  \subfigure[Destination Prediction] {
        \begin{minipage}[t]{1.0\linewidth}
        \centering
        \includegraphics[width=0.475\linewidth]{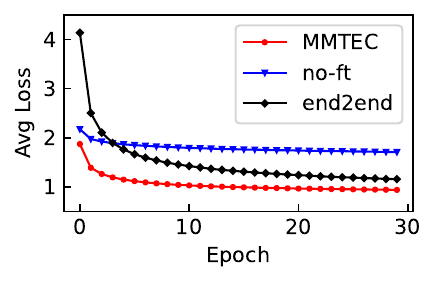}
	\includegraphics[width=0.485\linewidth]{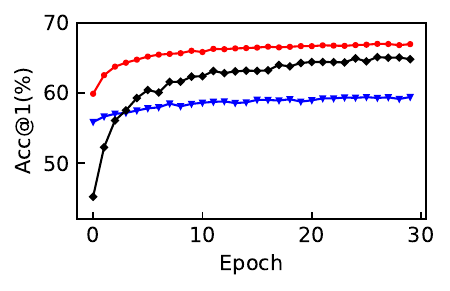}
        \end{minipage}
        \label{fig:epoch-losses-destination-prediction}
    }
    \subfigure[\alert{Travel Time Estimation}] {
        \begin{minipage}[t]{1.0\linewidth}
        \centering
        \includegraphics[width=0.48\linewidth]{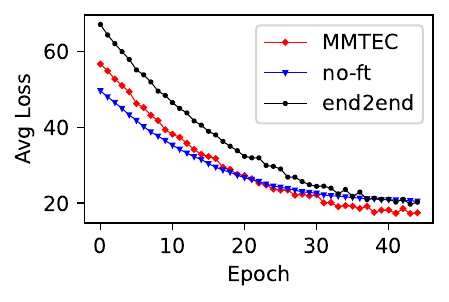}
	\includegraphics[width=0.48\linewidth]{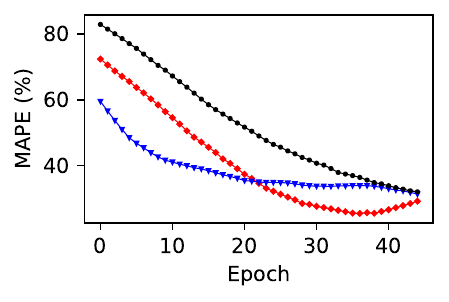}
        \end{minipage}
        \label{fig:epoch-losses-travel-time-estimation}
    }
\caption{Average loss value and validation metrics through the training processes of different variances.}
\label{fig:epoch-losses-acc}
\end{figure}

\subsubsection{Effectiveness of Pre-training}

We evaluate the effectiveness of the pre-training by comparing the complete MMTEC method with two variants of MMTEC. 
\begin{enumerate}
    \item \textit{no-ft}: not implementing fine-tuning. After pre-training, the parameters of the trajectory encoders are fixed and do not participate in the backward propagation. Only the parameters of the downstream predictor are trained in downstream tasks.
    \item \textit{end2end}: train the model in end-to-end manner. The parameters of the trajectory encoders are initialized randomly and are trained directly with supervision from downstream tasks.
\end{enumerate}

\alert{
The average loss values and validation metrics for destination prediction and travel time estimation on the Xian dataset are presented in Figure~\ref{fig:epoch-losses-acc}.
They show that the pre-training and fine-tuning approach facilitates accelerated convergence, signifying improved generalization and swifter progress in downstream tasks following the pre-training phase.
}

Pre-training without fine-tuning does not reach the performance of the fine-tuned and the end-to-end trained methods. Nevertheless, it still achieves satisfactory results, indicating that the pre-training embeddings capture general features of trajectories that are independent of task-specific supervision.

\subsubsection{Effectiveness of the Pretext Task}\label{sec:effectiveness-of-the-pretext-task}
To evaluate the effectiveness of the pretext task in our proposed method, we compare the trajectory embeddings learned by the complete MMTEC method with the embeddings learned when MMTEC uses the following pretext tasks for training:
\begin{enumerate}
    \item \textit{generative}: recover the original trajectory sequence given the embeddings. Here, we utilize two Transformer decoders to pair with the discrete and continuous trajectory encoders for trajectory recovery in an auto-regressive manner. The two encoders are trained separately.
    \item \textit{contrastive}: discriminating the positive sample from a set of negative samples given the target. 
    For each trajectory in a mini-batch, we regard its continuous spatio-temporal embedding as the target, its semantic embedding as the positive sample, and the semantic embeddings of other trajectories in the batch as negative samples. InfoNCE~\cite{oord2018representation} is chosen as the loss function.
\end{enumerate}

Their performance on various downstream tasks is shown in Tables~\ref{table:chengdu-overall-result} and \ref{table:xian-overall-result} and in Figure~\ref{fig:radar}.
We observe that when trained using the generative pretext task, the embeddings are biased towards point-level tasks such as travel time estimation and destination prediction. Conversely, when trained using the contrastive pretext task, the embeddings are biased towards sequence-level tasks such as similar trajectory search. Our full method, trained using the proposed MMTEC pretext task, possess consistent performance across different downstream tasks.

\subsubsection{Impact of Hyper-parameters} \label{sec:impact-of-hyper-parameters}

\begin{figure}[t]
  \centering
  \subfigure[Impact of embedding dimension $d$] {
    \begin{minipage}[t]{1.0\linewidth}
			\centering
			\includegraphics[width=0.48\linewidth]{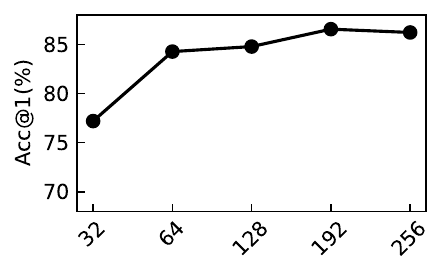}
			\includegraphics[width=0.48\linewidth]{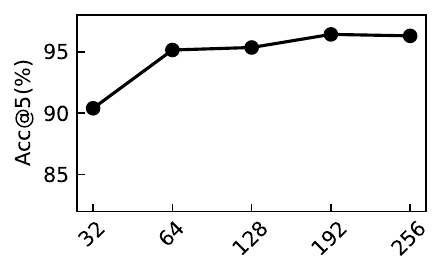}
		\end{minipage}
  \label{fig:parameter-d}
  }
  \subfigure[Impact of batch size $N$] {
    \begin{minipage}[t]{1.0\linewidth}
			\centering
			\includegraphics[width=0.48\linewidth]{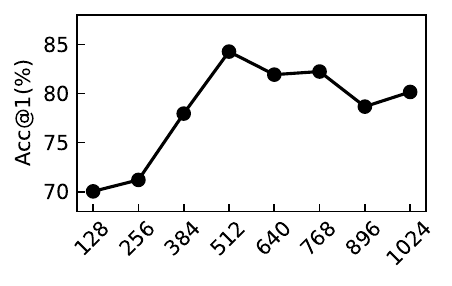}
			\includegraphics[width=0.48\linewidth]{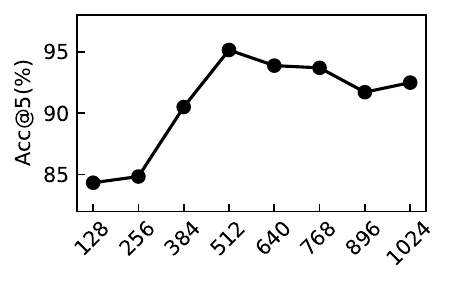}
		\end{minipage}
  \label{fig:parameter-N}
  }
  \subfigure[Impact of squared error upper bound $\epsilon^2$] {
    \begin{minipage}[t]{1.0\linewidth}
			\centering
			\includegraphics[width=0.48\linewidth]{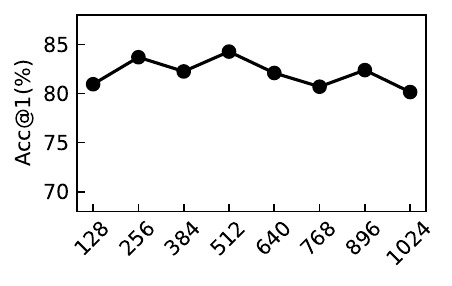}
			\includegraphics[width=0.48\linewidth]{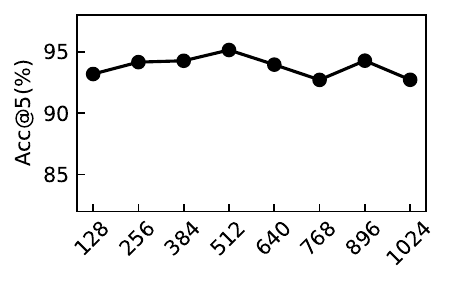}
		\end{minipage}
  \label{fig:parameter-eps}
  }
  \subfigure[Impact of anchor length $L_{A_1}$] {
    \begin{minipage}[t]{1.0\linewidth}
			\centering
			\includegraphics[width=0.48\linewidth]{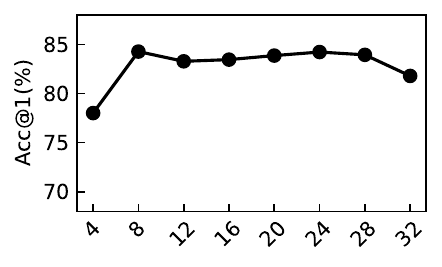}
			\includegraphics[width=0.48\linewidth]{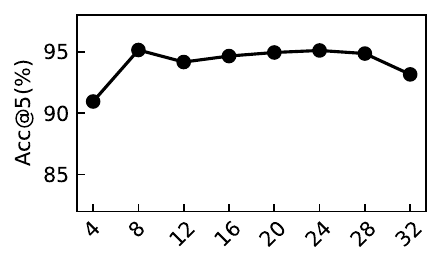}
		\end{minipage}
  \label{fig:parameter-LA}
  }
  \subfigure[Impact of Taylor expansion order $K$] {
    \begin{minipage}[t]{1.0\linewidth}
			\centering
			\includegraphics[width=0.48\linewidth]{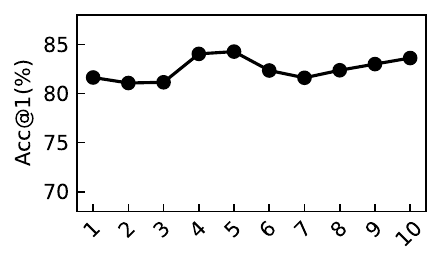}
			\includegraphics[width=0.48\linewidth]{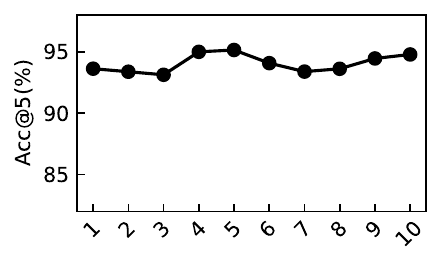}
		\end{minipage}
  \label{fig:parameter-K}
  }
	\caption{Impact of hyper-parameters for the similar trajectory search task on the Chengdu dataset.}
	\label{fig:hyper-parameters}
\end{figure}

We study the effects of the hyper-parameters listed in Table~\ref{table:hyper-parameters} on the test set of the Chengdu dataset. 
We use the Acc@$1$ and Acc@$5$ metrics of the similar trajectory search task, since the task does not involve fine-tuning and can indicate the quality of the learned trajectory embeddings directly. The results are presented in Figure~\ref{fig:hyper-parameters}, and we make the following observations:
\begin{enumerate}
    \item As illustrated in Figure~\ref{fig:parameter-d}, increasing the dimensionality of embeddings generally improves their performance. 
    However, beyond $d=64$, the accuracy improvement is negligible, while the computation and memory requirements increase significantly. Therefore, we set $d=64$ to balance performance and efficiency.
    \item Both the batch size $N$ and the squared error upper bound $\epsilon^2$ affect the pre-training loss function's coefficients (Equation~\ref{eq:mmtec-loss-function}). As shown in Figure~\ref{fig:parameter-N} and Figure~\ref{fig:parameter-eps}, their optimal value is $512$ in both cases.
    \item The anchor length $L_{A_1}$ of the first $\mathrm{IA}$ layer in the discrete trajectory encoder has an optimal value of $8$, as seen in Figure~\ref{fig:parameter-LA}. 
    A smaller length decreases model complexity, making it challenging to extract correlations from trajectories, while a longer length increases the model capacity, leading to overfitting.
    \item The Taylor expansion order $K$ determines the accuracy of the estimation of Equation~\ref{eq:prior-coding-length-function}. We set $K=5$ since it achieves good results.
\end{enumerate}

\subsubsection{Ablation Study}
To evaluate the effectiveness of different components of the proposed method, we compare the performance of learned embeddings obtained by the complete version and the following variants of the method:
\begin{enumerate}
    \item \textit{only-sem}: only use the semantic embeddings of trajectories in downstream tasks.
    \item \textit{only-con}: only use the continuous spatio-temporal embeddings of trajectories in downstream tasks.
    \item \textit{no-time}: remove the timestamp feature from both trajectory encoders.
    \item \textit{no-spatial}: remove the spatial coordinate features from both trajectory encoders.
    \item \textit{dis-no-time}: remove the timestamp feature from the discrete trajectory encoder.
    \item \textit{dis-no-spatial}: remove the spatial coordinate features from the discrete trajectory encoder.
    \item \textit{con-no-time}: remove the timestamp feature from the continuous trajectory encoder.
    \item \textit{con-no-spatial}: remove the spatial coordinate features from the continuous trajectory encoder.
    \item \textit{Dis-Trans}: replace the discrete trajectory encoder with a Transformer encoder.
    \item \textit{Con-Trans}: replace the continuous trajectory encoder with a Transformer encoder that takes the trajectory sequence $\mathcal T$ as input.
\end{enumerate}

\begin{table}[t]
    \centering
    \caption{Performance comparison of different variances on the similar trajectory search task, Chengdu dataset.}
    \label{table:ablation-study}
    \begin{threeparttable}
    \begin{tabular}{c|cc}
    \toprule
    Metric & Acc@1(\%) & Acc@5(\%) \\
    \midrule
    only-sem & 64.84$\pm$11.53 & 77.93$\pm$13.62 \\
    only-con & 59.36$\pm$4.67 & 71.23$\pm$5.32 \\
    \midrule
    no-time & 67.38$\pm$5.28 & 78.14$\pm$7.50 \\
    no-spatial & 53.10$\pm$0.51 & 64.23$\pm$1.04 \\
    dis-no-time & 70.72$\pm$0.49 & 82.53$\pm$0.91 \\
    dis-no-spatial & 72.71$\pm$2.14 & 83.93$\pm$2.85 \\
    con-no-time & 69.69$\pm$0.57 & 80.24$\pm$0.83 \\
    con-no-spatial & 62.79$\pm$0.59 & 73.56$\pm$0.91 \\
    \midrule
    Dis-Trans & \underline{82.86$\pm$2.57} & \underline{94.36$\pm$1.39} \\
    Con-Trans & 74.97$\pm$3.55 & 88.75$\pm$1.91 \\
    \textbf{MMTEC} & \textbf{84.27$\pm$3.02} & \textbf{95.15$\pm$1.63} \\
    \bottomrule
    \end{tabular}
    \begin{tablenotes}\footnotesize
        \item[]{\textbf{Bold} denotes the best result, 
        \underline{underline} denotes the second best result.}
    \end{tablenotes}
    \end{threeparttable}
\end{table}

For the reason stated in Section~\ref{sec:impact-of-hyper-parameters}, we compare the performance of these variants at the similar trajectory search task on the Chengdu dataset. The results, shown in Table~\ref{table:ablation-study}, lead to the following observations:
\begin{enumerate}
    \item Both the semantic and the continuous spatio-temporal embeddings are useful in downstream tasks. Using only one of them at similar trajectory search fails to match the accuracy of the full method. This indicates that both types of embeddings capture important and complementary information about the trajectories.
    \item The spatial and temporal features are essential for both trajectory encoders. Removing either  of these significantly degrades the accuracy. Since the two encoders are trained together during pre-training, even removing features from one encoder impacts the performance of the learned embeddings substantially.
    \item Replacing the proposed discrete trajectory encoder with a vanilla Transformer encoder degrades the accuracy somewhat. This suggests that the proposed encoder is capable of effectively and efficiently handling sequences of map-matched trajectories better than the Transformer. 
    Additionally, we observe a decrease in performance when replacing the continuous trajectory encoder with a Transformer encoder, as Transformers do not inherently model spatio-temporal correlations continuously.
\end{enumerate}

\section{Conclusion}\label{sec:conclusion}
We propose MMTEC, a novel pre-training trajectory embedding method that learns general trajectory embeddings that generalize to a wide range of downstream trajectory mining tasks, improving their performance. 
We propose a discrete trajectory encoder and a continuous trajectory encoder to extract semantic travel information and continuous spatio-temporal information from trajectories, respectively, improving the comprehensiveness of the learned trajectory embeddings.
The experimental study offers evidence of the improved effectiveness and performance over existing methods of the proposed method at pre-training high-quality trajectory embeddings.

\ifCLASSOPTIONcompsoc
  \section*{Acknowledgments}
\else
  \section*{Acknowledgment}
\fi

This work was supported by the Fundamental Research Funds for the Central Universities (Grant No. 2021YJS030).

\ifCLASSOPTIONcaptionsoff
  \newpage
\fi

\bibliographystyle{IEEEtran}
\bibliography{reference}

\begin{IEEEbiography}[{\includegraphics[width=1in,height=1.25in,clip,keepaspectratio]{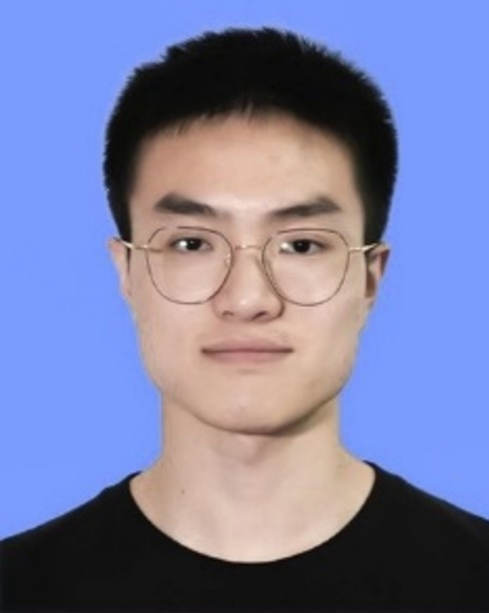}}]{Yan Lin} received the B.S. degree in computer science from Beijing Jiaotong University, Beijing, China, in 2019.

He is currently working toward the Ph.D. degree in the School of Computer and Information Technology, Beijing Jiaotong University. His research interests include spatio-temporal data mining and graph neural networks.
\end{IEEEbiography}

\begin{IEEEbiography}[{\includegraphics[width=1in,height=1.25in,clip,keepaspectratio]{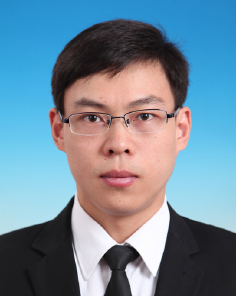}}]{Huaiyu Wan} received the Ph.D. degree in computer science and technology from Beijing Jiaotong University, Beijing, China, in 2012.

He is a Professor with the School of Computer and Information Technology, Beijing Jiaotong University. His current research interests focus on spatio-temporal data mining, social network mining, information extraction, and knowledge graph.
\end{IEEEbiography}

\begin{IEEEbiography}[{\includegraphics[width=1in,height=1.25in,clip,keepaspectratio]{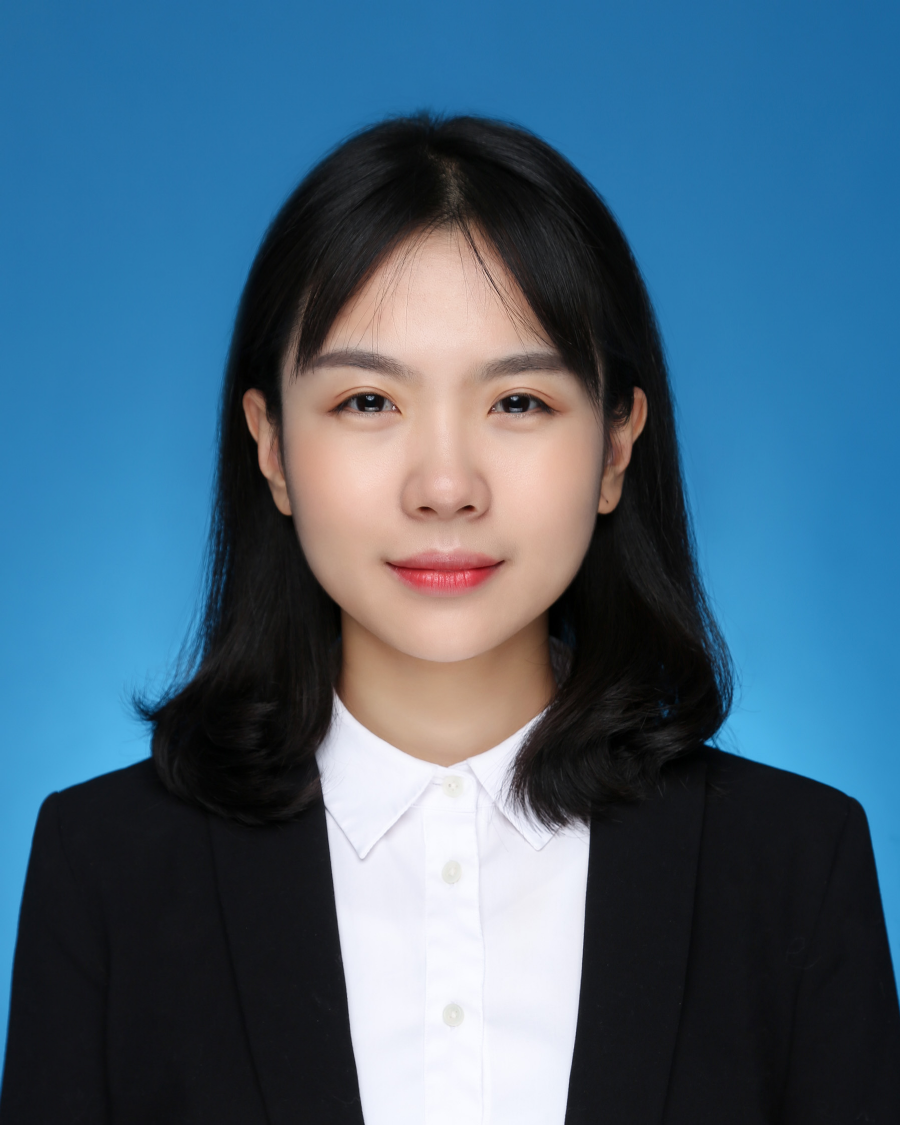}}]{Shengnan Guo} received the B.S. degree in computer science from Beijing Jiaotong University, Beijing, China, in 2015.

She is currently working toward the Ph.D. degree in the School of Computer and Information Technology, Beijing Jiaotong University. Her research interests focus on the area of deep learning and spatio-temporal data mining.
\end{IEEEbiography}

\begin{IEEEbiography}[{\includegraphics[width=1in,height=1.25in,clip,keepaspectratio]{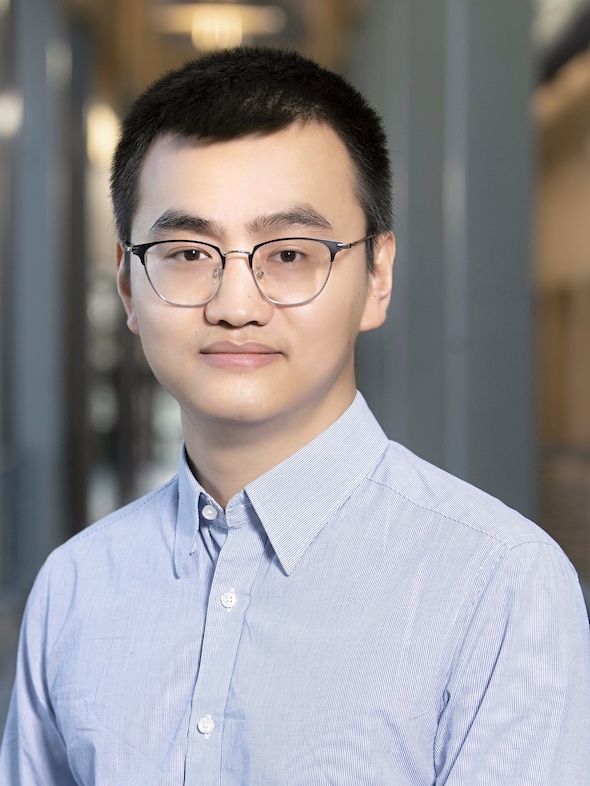}}]{Jilin Hu} received the Ph.D. degree in computer science from Aalborg University, Aalborg, Denmark in 2019.

He is an Associate Professor at the Department of Computer Science, Aalborg University. 
His research interests include spatio-temporal data analytics and transportation data mining.
\end{IEEEbiography}

\begin{IEEEbiography}[{\includegraphics[width=1in,height=1.25in,clip,keepaspectratio]{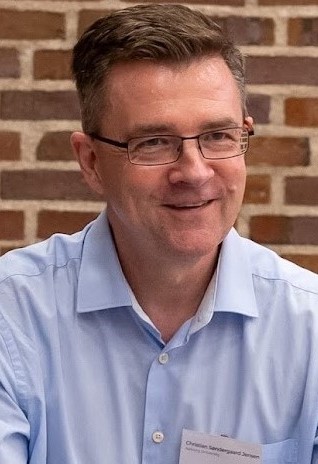}}]{Christian S. Jensen} received the Ph.D. degree from Aalborg University in 1991 after 2 1/2 years of study at University of Maryland, and he received the Dr.Techn. degree from Aalborg University in 2000.

He is a Professor at the Department of Computer Science, Aalborg University. 
His research concerns primarily temporal and spatio-temporal data management and analytics, including indexing and query processing, data mining, and machine learning.
\end{IEEEbiography}

\begin{IEEEbiography}[{\includegraphics[width=1in,height=1.25in,clip,keepaspectratio]{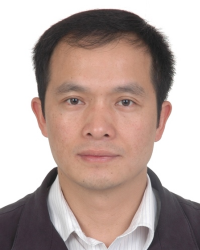}}]{Youfang Lin} received the Ph.D. degree in signal and information processing  from Beijing Jiaotong University, Beijing, China, in 2003.

He is a Professor with the School of Computer and Information Technology, Beijing Jiaotong University. His main fields of expertise and current research interests include big data technology, intelligent systems, complex networks, and traffic data mining.
\end{IEEEbiography}


\end{document}